%% file: main.tex
\documentclass{article}

\PassOptionsToPackage{numbers, sort&compress}{natbib}

\usepackage[main,final]{neurips_2026}
\usepackage{kotex}

\usepackage[utf8]{inputenc} 
\usepackage[T1]{fontenc}    
\usepackage{hyperref}       
\usepackage{url}            
\usepackage{booktabs}       
\usepackage{amsfonts}       
\usepackage{nicefrac}       
\usepackage{microtype}      
\usepackage{xcolor}         
\newcommand{\appref}[1]{\hyperref[#1]{Appendix~\ref*{#1}}}
\usepackage{algorithm}
\usepackage{algpseudocode}
\usepackage{etoolbox}
\hypersetup{hidelinks}

\usepackage{amsmath}
\usepackage{enumitem}
\usepackage{multirow}
\usepackage{graphicx}
\usepackage{tabularx}
\usepackage{adjustbox} 

\newcolumntype{C}{>{\centering\arraybackslash}X}

\title{
Baba in Wonderland: Online Self-Supervised Dynamics Discovery for Executable World Models}

\author{SeungWon Seo\thanks{These authors contributed equally to this work.}\\
  Korea University\\
  Seoul, Republic of Korea \\
  \texttt{ssw03270@korea.ac.kr}
  \And
  DongHeun Han\footnotemark[1]\\
  Kyung Hee University\\
  Yongin, Republic of Korea \\
  \texttt{hand32@khu.ac.kr}
  \AND
  SeongRae Noh\\
  Korea University\\
  Seoul, Republic of Korea \\
  \texttt{rhosunr99@korea.ac.kr}
  \And
  HyeongYeop Kang\thanks{Corresponding author.}\\
  Korea University\\
  Seoul, Republic of Korea \\
  \texttt{siamiz\_hkang@korea.ac.kr}
}

\begin{document}

\maketitle

\begin{abstract}
Executable world models can be read, edited, executed, and reused for planning, but only if the program captures the environment's transition law rather than semantic shortcuts in its surface vocabulary.
We study online executable world-model learning under prior misalignment, where an agent must induce state-dependent dynamics from interaction evidence alone, without rule descriptions, reward signals, or trustworthy lexical priors.
We introduce \textbf{Alice}, a closed-loop system that treats failed candidate updates as structural signal: when a candidate explains a new transition but loses previously explained ones, the preservation conflict reveals dynamics that the current program had conflated.
Alice refines these conflicts into hypothesis classes that both provide a few broken examples, one per affected class, for update and guide frontier exploration toward transitions that are novel and underrepresented with respect to the current program.
We evaluate Alice on \textit{Baba in Wonderland}, a prior-misaligned variant of \textit{Baba Is You} that preserves simulator dynamics while replacing semantically meaningful rule-property labels with unrelated words.
Experiments show that Alice substantially improves executable world-model learning under prior misalignment, and five-run ablations support overall improvement trends as class structure and exploration signals are added.
Project page: \url{https://ssw03270.github.io/BABAinWonderland_NeurIPS-26_Page/}.
\end{abstract}

\begin{figure*}[t]
\centering
\includegraphics[width=\textwidth]{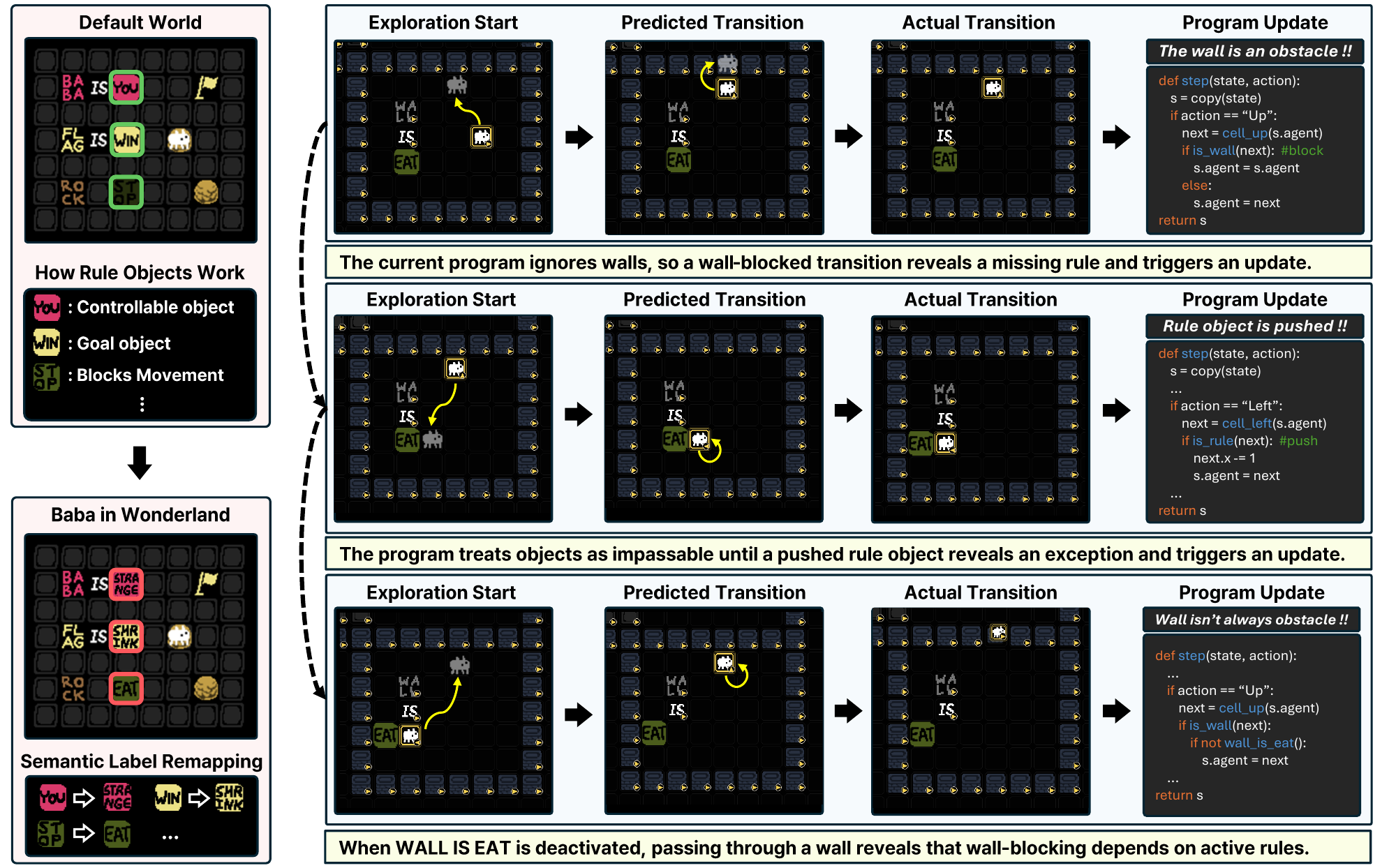}
\caption{
\textbf{Alice} in \textit{Baba in Wonderland}. 
Left: simulator dynamics are preserved while rule-property labels are semantically remapped, removing lexical shortcuts. 
Right: exploration exposes prediction errors that drive successive executable world-model revisions.
}
\label{fig:alice_teaser}
\end{figure*}

\section{Introduction}
\label{sec:intro}

Executable world models promise dynamics models that can be read, edited, executed, and reused for planning without post-hoc interpretation~\citep{tang2024worldcoder,lehrach2025code,dainese2024generating,piriyakulkij2025poe}. This promise, however, depends on whether the program captures the environment's transition law. When a language model synthesizes code in an environment whose surface vocabulary aligns with pretrained semantic priors, much of this inductive burden is supplied for free. When it does not, the same vocabulary becomes an active source of error~\citep{zhou2025wall,tang2024mars,xu2026code}, and the editable program, however clean its source, encodes the wrong dynamics.

Recent progress on executable and programmatic world models has largely operated on the favorable side of this divide, relying on curated transition traces, natural-language rule descriptions, or vocabularies whose surface meanings match the dynamics they label~\citep{tang2024worldcoder,lehrach2025code,dainese2024generating,piriyakulkij2025poe}. A separate line confronts regimes without such scaffolds, but typically reasons inside the Large Language Model (LLM) rather than synthesizing a persistent program~\citep{zhou2025wall,tang2024mars}. Their intersection -- online executable world-model construction without trustworthy surface vocabulary -- has received little direct study, yet this is where executable models are most useful: novel domains, opaque simulators, and unfamiliar terminology. The agent cannot be handed the rules; it must induce them from interaction, and an opaque neural model gives up much of the reason to seek executable form.

We instantiate this regime with \textbf{Baba in Wonderland}, a controlled prior-misaligned variant of \textit{Baba Is You}\footnote{\href{https://store.steampowered.com/app/736260/Baba_Is_You/}{Steam page for Baba Is You}.} The original game already exhibits state-dependent dynamics: rule sentences live inside the state, so the same action can produce different outcomes as rule blocks are rearranged. Yet labels such as \texttt{STOP}, \texttt{MOVE}, and \texttt{WIN} still provide semantic shortcuts. Baba in Wonderland preserves the simulator dynamics, action space, and rule-parsing mechanics, but replaces these labels with semantically unrelated words and provides no dictionary. A remapped token remains a stable identifier for the same role, but its surface form no longer hints at how that role behaves; successful modeling must be grounded in transition evidence rather than lexical priors.

Building an executable world model online in this regime exposes two coupled bottlenecks. The first is an \textit{update bottleneck}: each unexplained transition demands a code revision that explains the new transition without losing previously explained ones. Providing the full history in the update prompt is impractical because the dataset grows while the LLM context window does not~\citep{ma2024agentboard,wu2025resum,liu2024lost}. Updating from a small subset is context-efficient~\citep{xia2023keep} but risks brittle revisions that fit the target while silently regressing on rare or structurally distinct cases~\citep{zhou2023patchzero}. The question is therefore which subset can expose distinctions that the program itself has not yet made.

The second is an \textit{exploration bottleneck}. Improving the model requires transitions that expose gaps in its current hypotheses, but no external reward, rule description, or pretrained prior identifies which transitions those are~\citep{jin2020reward,tang2024mars,zhou2025wall}. Without such a signal, exploration either spends budget on transitions the current program already explains or traverses the state graph without knowing which transitions matter. What is needed is a notion of ``novelty with respect to the program being built", derived from interaction rather than rewards or priors.

Our central observation is that both bottlenecks share a hidden source of signal: \textit{the program's own update failures}. When a candidate revision explains a new transition but loses previously explained ones, the lost transitions are evidence that the earlier program had conflated distinct dynamics under a single branch of code. The preservation conflict reveals a latent partition of accumulated evidence into hypothesis classes that respond differently to the proposed change. Once explicit, this partition also defines what is underrepresented relative to the current program, turning a rejection event into the substrate for both compact update evidence and program-relative exploration.

We realize this idea in \textbf{Alice}, a closed-loop system that maintains a single persistent executable program while alternating LLM-based updates with update-guided exploration. Rejected candidate updates refine accumulated transitions into hypothesis classes, which serve two roles: they supply a few broken examples, one per affected class, for update, and they shape an embedding space in which a frontier explorer scores candidates by particle-entropy novelty and class-rarity coverage. Across Baba in Wonderland and the original Default World, Alice learns substantially more accurate executable world models than the closest prior interaction-driven baseline, with the gap widening sharply under prior misalignment; five-run ablations support overall improvement trends as class structure and exploration signals are added. Alice is a general method for online executable world-model learning, not a method specific to \textit{Baba Is You}. We use \textit{Baba in Wonderland} as a controlled probe to assess dynamics discovery from transition evidence while reducing reliance on pretrained knowledge of label meanings. We also evaluate Alice in the more realistic household setting of ALFWorld~\citep{shridhar2020alfworld}, extending the evaluation beyond \textit{Baba Is You} (\autoref{tab:alfworld_results}). \autoref{fig:alice_teaser} illustrates the setting and update loop.

We summarize our contributions as follows:
\begin{itemize}
    \item \textbf{A regime and a benchmark.} We introduce \textit{Baba in Wonderland}, a controlled variant of \textit{Baba Is You} in which simulator dynamics are preserved but surface labels are remapped to semantically unrelated words. As a controlled probe, the benchmark isolates online executable world-model learning in a regime where lexical priors offer no shortcut.
    \item \textbf{Failed updates as structural signal.} We show that rejected candidate program updates expose a latent partition of accumulated transitions into hypothesis classes, refined monotonically as more updates are attempted. This turns an event normally discarded as a rejection into a reusable artifact for organizing experience.
    \item \textbf{A closed loop that uses this signal twice.} The same hypothesis classes supply a few broken examples, one per affected class, for the next program update and shape an embedding in which a frontier explorer scores candidates by novelty and class rarity. One signal therefore drives both program updates and data collection.
    \item \textbf{Empirical evidence under prior misalignment.} On Baba in Wonderland, Alice substantially outperforms the closest interaction-driven executable world-model baseline; five-run ablations assess overall trends as class structure and exploration signals are added.
\end{itemize}

\section{Related Work}

\paragraph{Learning executable world models.}
World-model learning is often studied through neural latent dynamics models~\citep{ha2018world,hafner2019learning,hafner2023mastering,hansen2023td}, and recent work also uses LLMs as implicit simulators or reasoning backends for planning~\citep{wang2024can,huang2022language,hao2023reasoning,seo2025reveca,seo2026assumptions}.
These models can be powerful, but their dynamics are not preserved as a persistent executable artifact that can be inspected, locally updated, and reused as source code.
This distinction matters in environments where pretrained semantic priors conflict with the transition law~\citep{zhou2025wall,tang2024mars,xu2026code}: language-space reasoning can follow surface meanings even when those meanings are misleading or when dynamics vary with state.

The closest line of work asks LLMs to externalize environment knowledge as executable code or structured programmatic world models, including reusable procedures for embodied control~\citep{liang2023code,wang2023voyager}, interaction-driven code world models~\citep{tang2024worldcoder}, search-guided synthesis~\citep{dainese2024generating}, natural-language-rule-to-code game models~\citep{lehrach2025code}, and compositional code-based modeling~\citep{piriyakulkij2025poe}.
Our work shares the goal of learning an executable world model, but studies a stricter regime: the agent must maintain a persistent program online from interaction evidence alone, while surface vocabulary is prior-misaligned and the effective transition law changes with the state.
Baba Is AI~\citep{cloos2024baba} and Baba is Y'all 2.0~\citep{charity2022baba} study rule manipulation and collaborative level design, respectively. Our benchmark's novelty lies in the intervention rather than the game: it remaps only the surface labels while keeping the transition law and interaction protocol fixed, providing a controlled probe of dependence on lexical priors. Other studies probe post hoc whether foundation models have learned world structure~\citep{vafa2025has} or develop world models for agent interactions~\citep{yu2026reinforcement,xiao2602webworld}. Unlike these, Alice constructs a persistent executable model online and uses failed program updates to reveal distinctions between transition dynamics, reusing this evidence for both program refinement and exploration.

\paragraph{Self-supervised exploration for dynamics discovery.}
Online world-model learning also requires exploration: the agent must collect transitions that expose gaps in the current hypothesis rather than merely expand a dataset.
Task reward can direct data collection in model-based control~\citep{hafner2023mastering}, and solution-finding objectives can guide executable world-model agents~\citep{tang2024worldcoder}, but these objectives may be unavailable in reward-free settings~\citep{jin2020reward} or misaligned with dynamics discovery.
Intrinsic-motivation and reward-free methods instead use information gain or uncertainty~\citep{houthooft2016vime,kim2020active}, novelty~\citep{pathak2017curiosity,burda2018exploration}, disagreement~\citep{pathak2019self}, memory or directed exploration~\citep{ecoffet2019go,badia2020never}, and entropy or mutual information in behavior and transition spaces~\citep{eysenbach2018diversity,liu2021behavior,laskin2022unsupervised,wan2023deir}.

For executable world-model learning, however, generic diversity is not enough: exploration should be relative to the program being updated.
Some prior work makes exploration model-relative through learning progress~\citep{kim2020active} or anticipated model novelty~\citep{sekar2020planning}, but these signals are designed for neural predictors or downstream control adaptation.
Alice instead derives the exploration space from the executable update process itself.
Rejected candidate updates induce hypothesis classes over transition evidence, and the Explorer uses these classes to measure novelty and underrepresentation with respect to the current program's explanatory frontier.

\section{Problem Setting}
\label{sec:problem}
\subsection{Formal Setup}

We formulate online executable world-model learning in a deterministic Markov decision process (MDP). Let \(\mathcal{S}\) and \(\mathcal{A}\) denote discrete state and action spaces, and let \(T : \mathcal{S} \times \mathcal{A} \to \mathcal{S}\) denote the environment transition function. At each global interaction step \(t\), the agent observes a transition \(\tau_t = (s_t, a_t, s_{t+1})\) with \(s_{t+1} = T(s_t, a_t)\), and accumulates the dataset \(D_t = (\tau_1, \tau_2, \ldots, \tau_t)\).

Rather than learning a neural predictor, we represent the world model as a source-code program \(P\).
We interpret \(P\) as an executable hypothesis about \(T\), whose validity is tested only through observed transition evidence: given \((s,a)\), the program either returns a predicted next state \(\hat{s}'\) or fails to execute.
We say that \(P\) explains a transition \(\tau = (s,a,s')\) if it executes successfully on \((s,a)\) and outputs \(\hat{s}' = s'\), and we write \(E(P; D) = \{ \tau \in D \mid P \text{ explains } \tau \}\) for the explained subset of \(D\).

The agent operates under a finite global interaction budget \(H\), and the executable hypothesis is continually revised from the accumulated evidence in \(D_t\).
Let \(P_i\) denote the current accepted program, with \(i\) indexing accepted updates. When a newly observed transition \(\tau_{t+1}\) is not explained by \(P_i\), the agent proposes a candidate update \(Q\), which is accepted as \(P_{i+1}\) only if it explains \(\tau_{t+1}\) while preserving every previously explained transition: $E(P_i; D_t) \cup \{\tau_{t+1}\} \subseteq E(P_{i+1}; D_{t+1})$.
The system-level objective is to use the interaction budget $H$ to expand $E(P_i; D_t)$, and exploration contributes by selecting interactions likely to expose explanatory failures of $P_i$.

This is not an offline modeling problem over a fixed transition dataset.
In our setting, pre-collected traces and rule descriptions are unavailable, and task rewards do not specify which transitions improve the hypothesis.
Exploration therefore means collecting online evidence for refining the current executable hypothesis about \(T\), not maximizing external reward.

\subsection{Baba in Wonderland}
\label{sec:baba_in_wonderland_setting}

We instantiate this setting in \textit{Baba Is You}, a puzzle game in which rule sentences live inside the grid as movable text blocks. The environment is challenging on two fronts. Its state space is large and combinatorial, so unguided exploration cannot enumerate it within any realistic budget. More fundamentally, because rules are part of the state, the same action can produce different outcomes depending on which sentences are currently active. A correct hypothesis about $T$ must therefore capture how the currently active rule configuration determines transition outcomes --- not a fixed background dynamics, but a transition law that the agent's own actions can rewrite.

The original game, however, has one feature that obscures the difficulty we want to study. Its rule-property labels --- \texttt{STOP}, \texttt{MOVE}, \texttt{WIN} --- still carry their everyday meanings, so an LLM can often guess correct dynamics from surface vocabulary alone. This residual semantic shortcut conflates two distinct sources of success: induction from transition evidence, and lookup from pretrained priors. To separate them, we introduce \textit{Baba in Wonderland}, a minimal intervention that preserves the simulator dynamics, action space, and rule-parsing mechanics, and changes only the surface labels of \texttt{rule\_property} tokens, remapping them to semantically unrelated words; see \appref{sec:appendix_baba_is_you} for environment details and the full remapping.

The agent receives no dictionary and no natural-language explanation of the remapping. The role of each remapped token 
must therefore be inferred from transition evidence and from how the token participates in active rule sentences, rather than from pretrained lexical priors alone.

\section{Method}

Section~\ref{sec:problem} identified two coupled bottlenecks for online executable world-model learning: an \textit{update bottleneck}, in which each candidate revision must preserve every previously explained transition under a context budget that cannot hold all of $D_t$, and an \textit{exploration bottleneck}, in which no external signal indicates which transitions would expose gaps in the current hypothesis. 
We propose \textbf{Alice}, a closed-loop system that addresses both by treating the same artifact --- failed candidate updates --- as a structural signal: the conflicts that arise when a candidate gains $\tau_{t+1}$ but loses some previously explained transitions reveal which transitions belonged to distinct hypotheses all along. 

The system has two components that share this signal. The \textbf{Programmer} maintains the persistent program $P_i$ and uses preservation conflicts to organize $D_t$ into hypothesis classes that supply compact broken examples for the next update. The \textbf{Explorer} uses the same classes to score frontier candidates by both novelty and class underrepresentation, steering data collection toward the program's explanatory frontier. The two components operate in a closed loop, alternating program updates with update-guided exploration over the agent's interaction budget $H$. Algorithm~\ref{alg:alice} summarizes this loop.

\begin{figure*}[t]
\centering
\includegraphics[width=\linewidth]{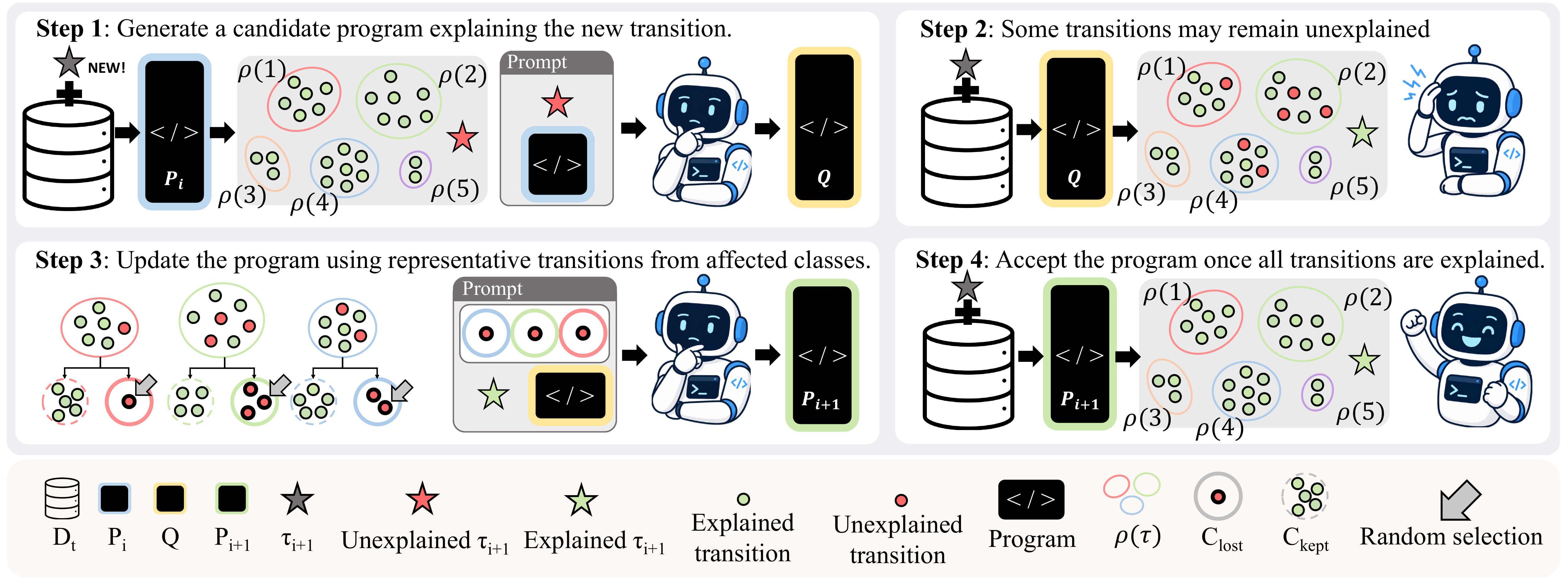}
\caption{Hypothesis-class refinement. Rejected target-explaining updates reveal lost transitions, split affected classes, and provide broken examples for later retries.}

\label{fig:class_overview}
\end{figure*}

\subsection{Refining Hypothesis Classes from Failed Updates}

When \(\tau_{t+1}\) is unexplained by $P_i$, the Programmer asks an LLM to propose a candidate update $Q$ that explains $\tau_{t+1}$ while preserving every transition in $E(P_i; D_t)$. Verifying preservation is straightforward: re-run $Q$ on every transition in $D_t$ and check that the explained set still contains $E(P_i; D_t)$. \textit{Generating} 
a candidate that will pass this check is harder, because the prompt must convey what those transitions actually look like, and neither $P_i$ nor $\tau_{t+1}$ alone is enough. $P_i$ is an executable hypothesis induced from $D_t$, not a lossless record of it: its code specifies how to compute next states without recording which transitions support each branch. The evidence must therefore come from $D_t$ itself --- but $D_t$ grows online while the prompt budget is fixed, so the Programmer cannot include it in full. The Programmer therefore selects a compact reference set $R_t \subseteq D_t$ to include in the prompt. The full inductive program-update prompt template is provided in \appref{sec:appendix_prompt_templates}. The question is which transitions should be in $R_t$.

The answer is that the LLM itself can produce this evidence whenever a candidate is rejected. Suppose $Q$ explains $\tau_{t+1}$ but fails to preserve some previously explained transitions. We call the set of lost transitions 
the \textit{broken examples} of $Q$,
\[
L_{\mathrm{pc}}(Q; P_i, D_t) = E(P_i; D_t) \setminus E(Q; D_t).
\]
These are transitions that $P_i$ explained, grouping them with whatever rule its code applies in the relevant branch, yet $Q$ now explains them differently from $\tau_{t+1}$. The merger inside $P_i$ was therefore hiding a distinction, and $L_{\mathrm{pc}}$ surfaces exactly the transitions where that distinction matters. Each rejected candidate that explains its target thus contributes one cut through $E(P_i; D_t)$, separating transitions that respond differently to the proposed update. The Programmer accumulates these cuts as a mutable collection of \textit{hypothesis classes} $\mathcal{C}$ over $E(P_i; D_t)$. Transitions remain in the same class until separated by a rejected candidate (\autoref{fig:class_overview}).

We initialize $\mathcal{C}$ from the accepted update sequence itself. Transitions that entered the explained set under the same bundle of hypotheses tend to provide redundant preservation evidence, so we group them by the earliest accepted version from which they have remained explained:
\[
\rho(\tau)=\min\{j \le i \mid \tau \in E(P_k;D_t)\ \text{for all } k=j,\ldots,i\}.
\]
For example, if $\tau$ is explained by $P_1$, lost by $P_2$, and explained by every version from $P_3$ onward, then $\rho(\tau) = 3$. Transitions sharing $\rho(\tau)$ form the \textit{root classes} of $\mathcal{C}$. Here, earlier program versions are evaluated retrospectively on the current dataset.

Root classes alone are coarse, however, because an early update can explain its target with a rule broader than the truth, sweeping in transitions that should later prove to obey something different. Rejected candidates expose such overgeneralizations at the class level. When $L_{\mathrm{pc}}(Q; P_i, D_t)$ covers only part of a class $C$, the candidate $Q$ separates $C$ into a preserved subset and a lost subset,
\[
C_{\mathrm{lost}} = C \cap L_{\mathrm{pc}}(Q;P_i,D_t), \qquad
C_{\mathrm{kept}} = C\setminus C_{\mathrm{lost}},
\]
giving evidence that $C$ had merged transitions responding differently to the new hypothesis. Whenever both subsets are 
nonempty, the Programmer replaces $C$ with the two refined classes and stores $Q$ as the split-inducing test that will route future transitions to the appropriate side. A refined class can be split again by later rejected candidates, making \(\mathcal{C}\) monotonically finer over retries.

To build the small reference set $R_t$, the Programmer groups the broken examples by their refined classes, chooses up to $n = 3$ affected classes, and samples $m = 1$ broken example from each. These few transitions, each witnessing a distinct preservation constraint, replace what would otherwise be random samples from $D_t$. The acceptance check itself remains over the full $E(P_i; D_t)$, so $R_t$ shapes only the update context, not the correctness criterion. After $Q$ is accepted, $\tau_{t+1}$ and later explained transitions inherit the existing refinements: they are routed first by $\rho$ into their root class, then by the stored split candidates into the appropriate refined class, so newly explained transitions immediately contribute evidence to the same structure that produced them. The same $\mathcal{C}$ also guides how the Explorer collects future 
transitions in~\autoref{sec:update_guided_exploration}.

\begin{figure*}[t]
\centering
\includegraphics[width=\linewidth]{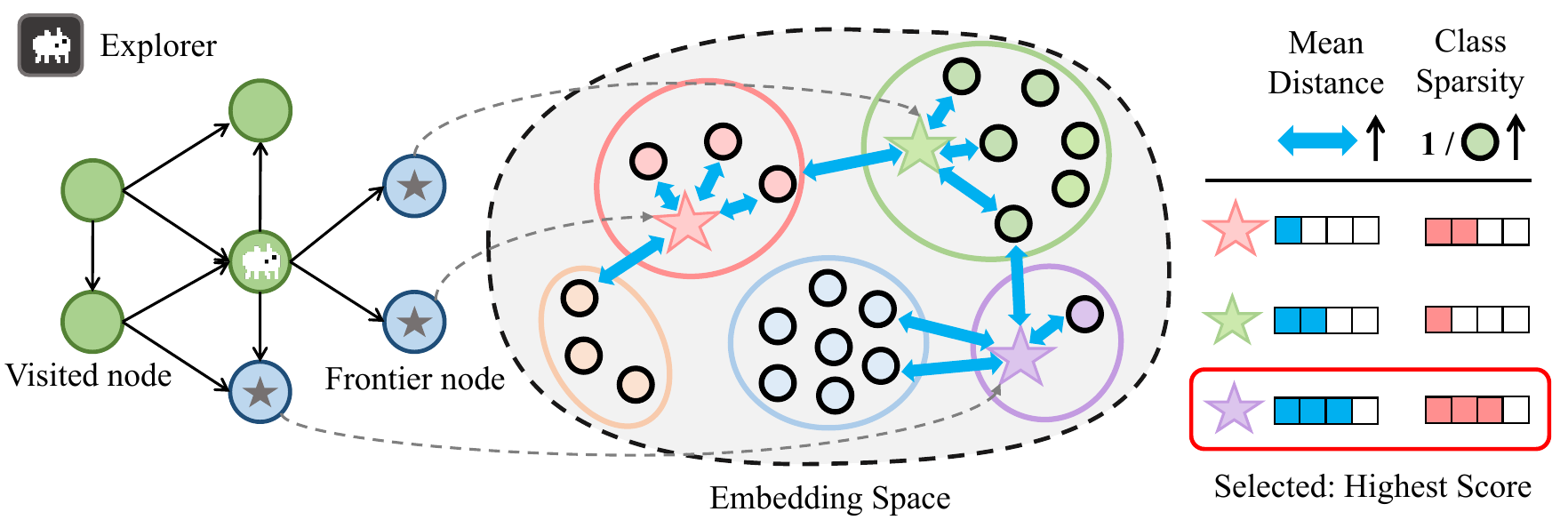}
\caption{Update-guided frontier scoring. The Explorer selects frontier candidates by combining embedding novelty with expected coverage of underrepresented hypothesis classes.}

\label{fig:exploration_overview}
\end{figure*}

\subsection{Update-Guided Exploration}
\label{sec:update_guided_exploration}
The Programmer turns transitions into program updates; the Explorer must turn the interaction budget $H$ into the informative transitions. A transition is informative for the next program update if two conditions hold: it should expose a failure of the current program $P_i$ so that an update is triggered, and ideally it should also induce a new hypothesis class so that $\mathcal{C}$ refines further. Neither property is observable before the candidate is executed. The Explorer therefore maintains a transition graph over reached states and scores each frontier candidate $x = (s, a)$, where $s$ is reached but unexpanded, using two pre-execution proxies: \textit{embedding novelty}, which favors candidates lying far from \(D_t\) in a learned state-action space, and \textit{expected class coverage}, which favors candidates likely to fall into hypothesis classes with little existing evidence.
Together, these proxies bias graph-frontier exploration toward the program's explanatory frontier.
\autoref{fig:exploration_overview} illustrates the procedure.

For embedding novelty to track the program's explanatory frontier rather than mere surface variation, candidates that fall in the same hypothesis class should map to nearby embeddings, and candidates from different classes should map to distant ones. The hypothesis classes $\mathcal{C}$ provide exactly the supervision needed for this geometry: the Explorer trains $h_\theta(x)$ with the supervised contrastive loss of Khosla et al.~\cite{khosla2020supervised}, pulling together embeddings within the same class and pushing apart embeddings across classes.
Supervision is limited to active classes \(\mathcal{C}_{\mathrm{act}}\subseteq\mathcal{C}\) with enough transitions for stable positives; low-count classes remain in \(\mathcal{C}\) but are excluded.

Given this embedding, novelty is measured by a $k$-nearest-neighbor particle-entropy score~\cite{liu2021behavior}, 
\[
r_h(x) = \log\!\left(1+\frac{1}{k}
\sum_{\tilde h\in \mathcal{N}_k(h_\theta(x);\mathcal{B}_h)}
\|h_\theta(x)-\tilde h\|_2
\right),
\]
where $\mathcal{B}_h$ is the embedding set induced by $D_t$ and $\mathcal{N}_k$ returns the $k = 32$ nearest neighbors. Candidates far from previously observed embeddings score high.

The second proxy, \textit{expected class coverage}, biases exploration toward classes with little existing evidence. Using the same active class set, each $C \in \mathcal{C}_{\mathrm{act}}$ contributes a class-count self-information $u(C) = -\log(|C|/N)$, where $N = \sum_{C \in \mathcal{C}_{\mathrm{act}}} |C|$ is the total count across active classes; rare classes have larger $u(C)$. Since the class label of an unexecuted candidate is itself unobservable, the Explorer estimates it from class prototypes --- up to 8 representative embeddings per class. A softmax over cosine similarities to prototype embeddings yields a pre-execution class distribution $q_\theta(C \mid h_\theta(x))$, and the expected class-coverage score is
\[
r_{\mathcal{C}}(x) = \sum_{C\in\mathcal{C}_{\mathrm{act}}}
q_\theta(C\mid h_\theta(x))\,u(C).
\]
The two proxies combine into the frontier score
\[
S_{\mathrm{frontier}}(x)
=
\lambda_h r_h(x)
+
\lambda_{\mathcal{C}} r_{\mathcal{C}}(x).
\]
with $\lambda_h = 1$ and $\lambda_{\mathcal{C}} = 0.35$ in the main online experiments; \appref{sec:appendix_revised_results} reports the sensitivity analysis. Each collection batch greedily selects the highest-scoring candidates from the current frontier under the interaction budget.

Executed transitions feed back into the loop. Explained transitions sharpen the class-structured evidence used to train $h_\theta$, while unexplained transitions become the Programmer's next explanatory failures. The single signal of failed updates therefore drives both halves of Alice: it shapes which broken examples the LLM sees in $R_t$, and it shapes which transitions the agent seeks through $S_{\mathrm{frontier}}$. \appref{sec:appendix_frontier_score} relates this score to an entropy--mutual-information objective.

\section{Experiments}
We evaluate Alice along two axes that jointly support the claims in~\autoref{sec:intro}: (i) end-to-end accuracy of the learned executable world model under prior misalignment, both for online interaction on regular levels and for held-out generalization to extra levels; and (ii) ablations assessing trends as components are added. Robustness to LLM backbones, additional diagnostic experiments, and qualitative examples are reported in~\appref{sec:appendix_additional_experiments} and~\appref{sec:appendix_qualitative_examples}.

\subsection{Experimental Setup}
\label{sec:exp-setup}

\paragraph{Settings.} We evaluate executable world-model learning on levels drawn from \textit{Baba Is You}, comprising $32$ regular levels and $8$ extra levels. 
Each level is instantiated under two label settings: \textit{Default World}, which uses the original \texttt{rule\_property} labels, and \textit{Baba in Wonderland}, which preserves the simulator dynamics but replaces the surface labels of \texttt{rule\_property} tokens with semantically unrelated labels (\autoref{tab:wonderland_remap}). 
The agent is given no dictionary and no description of the remapping.
Online experiments test learning from self-collected evidence: they collect interaction data on regular levels and evaluate on human-play solution transitions from the same levels. Offline experiments test generalization from a fixed dataset to unseen levels: they train on regular-level solution transitions and evaluate on held-out extra-level transitions, so their accuracies are not directly comparable with the online ones. Extra-level evaluation of the online programs is reported in \appref{sec:appendix_revised_results}. Additional protocol details appear in~\appref{sec:appendix_experimental_details}.

To evaluate a more realistic domain, we use ALFWorld's household tasks~\citep{shridhar2020alfworld} with a fully observable symbolic state/action interface following WorldCoder~\citep{tang2024worldcoder}. We train on all six task types without a curriculum and evaluate exact next-state prediction on unseen episodes. Each method receives up to 100,000 interaction steps before a program update, with the same 15-call update budget and 100-call run cap used in Baba. \appref{sec:appendix_alfworld} provides further protocol details.

\paragraph{Metrics.} 
\textit{All Acc.} is the fraction of evaluated transitions for which the predicted next state exactly matches the environment next state.
\textit{Balanced Acc.} applies the same exact-match criterion on a class-reduced evaluation set in which transitions are grouped by action and canonical state-difference signature, with one representative sampled per group; details appear in \appref{sec:appendix_heuristic_dynamics}.
Compile failures and runtime errors are counted as incorrect predictions. 
We additionally report \textit{Calls}, the cumulative number of LLM calls used during learning. 
For online runs each method is capped at \(100\) total LLM calls and \(15\) calls per single update iteration; offline runs are similarly capped at \(100\) total LLM calls.

\paragraph{Baselines.} For online experiments, we compare against \textbf{WorldCoder}~\citep{tang2024worldcoder}, the closest prior interaction-driven executable world-model learner. Like Alice, WorldCoder maintains executable code as its world model, but it does not organize failed program updates into hypothesis classes for later preservation or exploration.
For offline fixed-data experiments, we additionally compare against \textbf{GIF-MCTS}~\citep{dainese2024generating} and \textbf{CWM}~\citep{lehrach2025code}, which together cover search-based and population-based code world-model synthesis. All baselines use the same LLM backbone (GPT-5.4) and the same LLM-call caps as Alice; in the offline comparison Alice trains only on the fixed dataset and does not use its Explorer.
The main comparisons and ablations report mean $\pm$ standard deviation over five independent runs.

\subsection{Executable World-Model Learning}
\label{sec:exp-main}

\begin{table*}[t]
	\centering
	\small
    \setlength{\tabcolsep}{2pt}
    \caption{Online interaction results on regular levels. All Acc. and Balanced Acc. are exact one-step transition accuracies; Calls reports learning-time LLM calls. Five-run mean $\pm$ standard deviation.}

	\begin{tabular*}{\textwidth}{@{\hspace{4pt}\extracolsep{\fill}}lcccccc@{\hspace{4pt}}}
		\toprule
		& \multicolumn{3}{c}{Default World}
		& \multicolumn{3}{c}{Baba in Wonderland} \\
		\cmidrule(lr){2-4} \cmidrule(lr){5-7}
		Method
		& All Acc. $\uparrow$
		& Balanced Acc. $\uparrow$
		& Calls 
		& All Acc. $\uparrow$
		& Balanced Acc. $\uparrow$
		& Calls \\
		\midrule
		\textbf{WorldCoder} & $0.871\pm0.054$ & $0.727\pm0.160$ & $61.8\pm21.5$ & $0.542\pm0.113$ & $0.260\pm0.126$ & $57.6\pm5.5$ \\
		\textbf{Alice} & $0.988\pm0.008$ & $0.990\pm0.004$ & $19.8\pm6.6$ & $0.965\pm0.054$ & $0.951\pm0.068$ & $94.0\pm10.4$ \\
		\bottomrule
	\end{tabular*}
	\label{tab:main_transition}
\end{table*}

\begin{table*}[t]
	\centering
	\small
    \setlength{\tabcolsep}{2pt}
    \caption{Offline fixed-data results. Methods train on regular-level solution transitions and evaluate on held-out extra-level transitions. Five-run mean $\pm$ standard deviation.}
	\begin{tabular*}{\textwidth}{@{\hspace{4pt}\extracolsep{\fill}}lcccccc@{\hspace{4pt}}}
		\toprule
		& \multicolumn{3}{c}{Default World}
		& \multicolumn{3}{c}{Baba in Wonderland} \\
		\cmidrule(lr){2-4} \cmidrule(lr){5-7}
		Method
		& All Acc. $\uparrow$
		& Balanced Acc. $\uparrow$
		& Calls
		& All Acc. $\uparrow$
		& Balanced Acc. $\uparrow$
		& Calls \\
		\midrule
		\textbf{GIF-MCTS} & $0.853\pm0.046$ & $0.754\pm0.151$ & $100.0\pm0.0$ & $0.796\pm0.065$ & $0.716\pm0.061$ & $100.0\pm0.0$ \\
		\textbf{CWM} & $0.964\pm0.047$ & $0.934\pm0.058$ & $100.0\pm0.0$ & $0.845\pm0.051$ & $0.702\pm0.149$ & $100.0\pm0.0$ \\
		\textbf{Alice} & $1.000\pm0.000$ & $1.000\pm0.000$ & $21.4\pm1.5$ & $0.948\pm0.038$ & $0.920\pm0.047$ & $91.8\pm8.8$ \\
		\bottomrule
	\end{tabular*}
	\label{tab:offline_dataset_main}
\end{table*}

\paragraph{Online interaction.}
\autoref{tab:main_transition} reports online performance on regular levels in both label settings.
In \textit{Default World}, WorldCoder remains relatively strong but trails Alice in both All Acc. and Balanced Acc., reflecting that semantic priors help with many common transitions but can still miss rare or compositional dynamics.
The gap widens sharply in \textit{Baba in Wonderland}, where lexical priors are unreliable: WorldCoder's All Acc. drops by $0.329$ and its Balanced Acc. falls to $0.260$, while Alice retains $0.965$ All Acc. and $0.951$ Balanced Acc.

The WorldCoder call counts deserve attention. In \textit{Baba in Wonderland}, WorldCoder uses $57.6$ calls on average, partly because its update loop terminates after exhausting the per-update retry budget. Allowing it to continue to the same 100-call cap raises its All Acc. to $0.819\pm0.052$ and Balanced Acc. to $0.658\pm0.131$, still below Alice (\appref{sec:appendix_revised_results}). Alice addresses this failure mode by supplying the LLM with a few broken examples that target the constraints a naive update tends to violate.

\paragraph{Fixed-data comparison.}
\autoref{tab:offline_dataset_main} compares Alice with GIF-MCTS and CWM when all methods learn from the same fixed regular-level transitions and are evaluated on held-out extra-level transitions. In \textit{Default World}, semantically aligned surface labels make executable-rule synthesis easier for all methods, and Alice reaches perfect held-out accuracy with $21.4$ LLM calls on average. CWM is the strongest baseline in this setting, yet its
remaining Balanced Acc. gap indicates that high overall accuracy can still miss less frequent transition types. The comparison becomes more diagnostic in \textit{Baba in Wonderland}: because no method can collect additional transitions from the unseen extra levels, held-out accuracy depends on how effectively each method converts the fixed train evidence into executable dynamics without relying on lexical shortcuts. Here Alice improves over both GIF-MCTS and CWM at a weighted token cost between theirs (\appref{sec:appendix_cost}), with the largest margin in Balanced Acc., showing that the persistent update loop better preserves rare or easily conflated dynamics under prior misalignment.

\paragraph{World-model learning in ALFWorld.}
\label{sec:exp-alfworld}
\autoref{tab:alfworld_results} evaluates Alice on realistic household tasks. Alice achieves $0.962\pm0.024$ accuracy with $34.4\pm7.9$ calls, compared with WorldCoder's $0.702\pm0.077$ at the 100-call cap. Adding a six-task curriculum improves WorldCoder to $0.855\pm0.027$, indicating that task ordering explains part of the gap. Alice reaches higher accuracy without a curriculum, demonstrating applicability beyond rule-manipulation puzzles.

\begin{table}[t]
\centering\small
\setlength{\tabcolsep}{2pt}
\caption{Exact next-state prediction on unseen ALFWorld episodes. The third row adds a curriculum. Five-run mean $\pm$ standard deviation.}
\label{tab:alfworld_results}
\begin{tabular*}{\textwidth}{@{\hspace{4pt}\extracolsep{\fill}}lcc@{\hspace{4pt}}}
\toprule
Method & Accuracy $\uparrow$ & Calls \\
\midrule
\textbf{Alice} & $0.962\pm0.024$ & $34.4\pm7.9$ \\
\textbf{WorldCoder} & $0.702\pm0.077$ & $100.0\pm0.0$ \\
\textbf{WorldCoder + curriculum} & $0.855\pm0.027$ & $100.0\pm0.0$ \\
\bottomrule
\end{tabular*}
\end{table}

\subsection{Ablation Study Results}

\begin{table}[t]
\centering\small
\setlength{\tabcolsep}{2pt}
\caption{Online ablations on regular-level Wonderland solution transitions. Five-run mean $\pm$ standard deviation.}
\label{tab:main_ablations}
\begin{tabular*}{\textwidth}{@{\hspace{4pt}\extracolsep{\fill}}lccc@{\hspace{4pt}}}
\toprule
Setting & All Acc. $\uparrow$ & Balanced Acc. $\uparrow$ & Calls \\
\midrule
\textbf{Single-Class} & $0.890\pm0.053$ & $0.829\pm0.073$ & $84.8\pm20.9$ \\
\textbf{Root-Class} & $0.940\pm0.052$ & $0.906\pm0.071$ & $80.0\pm12.8$ \\
\textbf{BFS} & $0.901\pm0.071$ & $0.854\pm0.105$ & $86.8\pm18.6$ \\
\textbf{w/o} $r_h$ & $0.953\pm0.044$ & $0.924\pm0.053$ & $83.0\pm13.5$ \\
\textbf{w/o} $r_{\mathcal C}$ & $0.929\pm0.050$ & $0.890\pm0.062$ & $87.0\pm20.5$ \\
\textbf{Alice} & $0.965\pm0.054$ & $0.951\pm0.068$ & $94.0\pm10.4$ \\
\bottomrule
\end{tabular*}
\end{table}

\paragraph{Hypothesis classes.}
We ask how much structure is needed for the Programmer to turn broken examples into accepted program updates.
\textbf{Single-Class} collapses all explained transitions into one class, while \textbf{Root-Class} keeps only the root classes induced by \(\rho(\tau)\), before any refinement from rejected candidates.
Balanced accuracy rises from $0.829$ with Single-Class to $0.906$ with Root-Class and reaches $0.951$ with Alice (\autoref{tab:main_ablations}). Statistical tests support this overall increasing trend (\appref{sec:appendix_revised_results}). This pattern supports the role of failed updates in separating preservation constraints that remain conflated within coarse root classes. The resulting class-stratified \(R_t\) exposes these distinctions to the Programmer with a few broken examples.

\paragraph{Exploration objective.}
We ask whether the Explorer must actively shape which transition evidence is collected before each program update.
\textbf{BFS} selects frontier candidates in breadth-first order, removing program-relative frontier scoring; \textbf{w/o \(r_h\)} and \textbf{w/o \(r_C\)} remove the embedding-novelty and expected-class-coverage terms from \(S_{\text{frontier}}\), respectively.
Both score ablations have higher mean accuracy than BFS but lower than Alice. This pattern is consistent with complementary roles for the two score terms: \(r_h\) favors novelty in the learned state-action space, while \(r_C\) directs exploration toward underrepresented hypothesis classes. Their combination helps collect evidence that is both novel and relevant to weakly supported parts of the current program. See \appref{sec:appendix_revised_results} for significance tests. Learning curves and embedding visualizations provide additional analyses in \appref{sec:appendix_additional_experiments}.

\section{Conclusion}

We studied online executable world-model learning under prior misalignment, where state-dependent dynamics must be induced from interaction evidence rather than rule descriptions, rewards, or trustworthy lexical priors.
Alice maintains a persistent executable program and turns failed candidate updates into hypothesis classes that support both class-stratified program update and program-relative frontier exploration.
Through \textit{Baba in Wonderland}, we show that this closed loop can ground executable world-model construction in transition evidence even when surface labels are misleading. Experiments on ALFWorld demonstrate Alice's applicability to realistic household tasks beyond rule-manipulation puzzles.

However, our experiments are limited to deterministic, discrete, fully observable symbolic environments where transition explainability can be checked exactly by executing a program. Full-set checking also incurs increasing local computation as evidence accumulates, as detailed in \appref{sec:appendix_cost}.
Extending the idea to stochastic, continuous, or partially observable domains will require broader notions of explanation, preservation, and update-induced evidence.

\clearpage
\begin{ack}
This work was supported by the ICT Creative Consilience Program through the Institute of Information \& Communications Technology Planning \& Evaluation (IITP) grant funded by the Korea government (MSIT) (IITP-2026-RS-2020-II201819), and by the IITP grant funded by the Korea government (MSIT) (IITP-2026-RS-2026-25615490, AI Star Fellowship Support Program). The authors declare no competing interests.
\end{ack}

\bibliographystyle{plainnat}
\bibliography{reference}
\clearpage
\appendix

\section{Appendix}

\subsection{Baba Is You}
\label{sec:appendix_baba_is_you}

\paragraph{Game rules and dynamics.}
Baba Is You is a rule-manipulation puzzle game in which object behavior is determined by textual rules assembled directly in the grid world.
Rules become active when text blocks form valid horizontal or vertical rule sentences, and any change in that arrangement can activate, deactivate, or modify the effective dynamics of the environment.
Because these rule sentences can be created, broken, moved, or rearranged by the agent's actions, the currently active rule set is itself part of the environment state rather than a fixed background specification.
As a result, predicting the next state requires more than tracking the visible arrangement of objects: the same action can lead to different outcomes depending on which rules are currently active.
This makes the game a particularly challenging environment for world-model learning, because its transition law is state-dependent and symbolic rather than fixed and purely geometric.
In \textit{Default World}, for example, changing whether a sentence such as \texttt{WALL IS STOP} is active can immediately alter whether walls behave as obstacles, even when the surrounding object layout remains unchanged.
Baba Is You therefore provides a natural testbed for executable world-model learning in environments where accurate prediction depends on maintaining explicit hypotheses about mutable rules.

\paragraph{Vocabulary.}

\begin{table*}[h]
	\centering
	\small
\setlength{\tabcolsep}{2pt}
	\caption{Token vocabulary used in the \textit{Baba Is You} benchmark environment.}
	\begin{tabularx}{\textwidth}{@{\hspace{4pt}}lX@{\hspace{4pt}}}
		\toprule
		Type & Vocabulary \\
		\midrule
		\texttt{rule\_noun} & \texttt{ALGAE}, \texttt{BABA}, \texttt{BOG}, \texttt{BOLT}, \texttt{BRICK}, \texttt{BUBBLE}, \texttt{COG}, \texttt{CRAB}, \texttt{DOOR}, \texttt{FLAG}, \texttt{FLOWER}, \texttt{GRASS}, \texttt{HEDGE}, \texttt{ICE}, \texttt{JELLY}, \texttt{KEKE}, \texttt{KEY}, \texttt{LAVA}, \texttt{LOVE}, \texttt{PILLAR}, \texttt{PIPE}, \texttt{REED}, \texttt{ROCK}, \texttt{ROBOT}, \texttt{SKULL}, \texttt{STAR}, \texttt{TEXT}, \texttt{TILE}, \texttt{WALL}, \texttt{WATER} \\
		\texttt{rule\_operator} & \texttt{AND}, \texttt{IS} \\
		\texttt{rule\_property} & \texttt{DEFEAT}, \texttt{FLOAT}, \texttt{HOT}, \texttt{MELT}, \texttt{MOVE}, \texttt{OPEN}, \texttt{PUSH}, \texttt{SHUT}, \texttt{SINK}, \texttt{STOP}, \texttt{WIN}, \texttt{YOU} \\
		\texttt{world\_object} & \texttt{ALGAE}, \texttt{BABA}, \texttt{BOG}, \texttt{BOLT}, \texttt{BRICK}, \texttt{BUBBLE}, \texttt{COG}, \texttt{CRAB}, \texttt{DOOR}, \texttt{FLAG}, \texttt{FLOWER}, \texttt{GRASS}, \texttt{HEDGE}, \texttt{ICE}, \texttt{JELLY}, \texttt{KEKE}, \texttt{KEY}, \texttt{LAVA}, \texttt{LOVE}, \texttt{PILLAR}, \texttt{PIPE}, \texttt{REED}, \texttt{ROCK}, \texttt{ROBOT}, \texttt{SKULL}, \texttt{STAR}, \texttt{TILE}, \texttt{WALL}, \texttt{WATER} \\
		\bottomrule
	\end{tabularx}
	\label{tab:baba_token_vocab}
\end{table*}

\begin{table*}[h]
	\centering
	\small
\setlength{\tabcolsep}{2pt}
	\caption{Rule properties and their roles in the simulator used by the benchmark.}
	\begin{tabularx}{\textwidth}{@{\hspace{4pt}}lX@{\hspace{4pt}}}
		\toprule
		Property & Role \\
		\midrule
		\texttt{YOU} & Makes an object controllable during the agent-driven movement stage. \\
		\texttt{WIN} & Triggers success when a \texttt{YOU} object overlaps a winning object on the same float layer after interaction effects are resolved. \\
		\texttt{STOP} & Blocks entry and movement through the object. \\
		\texttt{PUSH} & Allows the object to be pushed by movers. \\
		\texttt{MOVE} & Causes the object to move automatically each turn, reversing direction if blocked. \\
		\texttt{DEFEAT} & Destroys overlapping \texttt{YOU} objects on the same float layer. \\
		\texttt{SINK} & Removes both itself and other overlapping same-layer objects. \\
		\texttt{HOT} & Destroys overlapping \texttt{MELT} objects on the same float layer. \\
		\texttt{MELT} & Is removed when overlapping a \texttt{HOT} object on the same float layer. \\
		\texttt{OPEN} & Cancels with \texttt{SHUT}, removing both on overlap. \\
		\texttt{SHUT} & Cancels with \texttt{OPEN}, removing both on overlap. \\
		\texttt{FLOAT} & Restricts many overlap-triggered interactions to objects on the same float layer. \\
		\bottomrule
	\end{tabularx}
	\label{tab:baba_property_roles}
\end{table*}

We use four token types throughout the benchmark: \texttt{rule\_noun}, \texttt{rule\_operator}, \texttt{rule\_property}, and \texttt{world\_object}.
\autoref{tab:baba_token_vocab} summarizes the vocabulary used by each type, and \autoref{tab:baba_property_roles} summarizes the rule properties and their roles in the simulator.
Rule nouns and world objects share most of their lexical stems, but they play different semantic roles: the former appear as text blocks in rule sentences, whereas the latter are instantiated entities in the grid world.
Among rule operators, \texttt{IS} links an object noun to a property or another noun, while \texttt{AND} allows conjunctions on either side of a rule, enabling forms such as \texttt{BABA AND ROCK IS YOU} and \texttt{BABA IS YOU AND PUSH}.
Crucially, rule properties are not merely labels.
They alter movement, collision, overlap, destruction, and goal conditions, and multiple active properties can interact within the same turn.
These interactions are also order-sensitive.
In the original game, their effects are resolved over an ordered update procedure rather than all at once.
Broadly, movement-related properties such as \texttt{YOU}, \texttt{MOVE}, and \texttt{PUSH} shape how objects are displaced, whereas overlap-triggered properties such as \texttt{DEFEAT}, \texttt{SINK}, \texttt{HOT}/\texttt{MELT}, \texttt{OPEN}/\texttt{SHUT}, and \texttt{WIN} are checked later in the turn.
\texttt{WIN} is evaluated especially late, so earlier destructive effects can prevent a win that would otherwise occur.
\texttt{FLOAT} further constrains many of these interactions by requiring the relevant objects to occupy the same float layer.
As a result, learning an accurate executable world model requires not only inferring which properties are currently active, but also capturing how their effects compose under the simulator's execution order.

\paragraph{State and action representation.}
We provide states to the LLM as structured JSON rather than as free-form natural-language descriptions.
Each state contains the top-level fields \texttt{grid\_size}, \texttt{step}, and \texttt{objects}.
\texttt{grid\_size} gives the width and height of the playable interior grid, after trimming the immutable outer border used by the underlying simulator.
The \texttt{step} field records episode-level status information and currently includes a Boolean \texttt{terminated} flag.
Each entry in \texttt{objects} is a dictionary with keys such as \texttt{type}, \texttt{word}, and \texttt{position}, and directional objects additionally include a \texttt{direction} field.
Positions are represented as zero-indexed \([x,y]\) coordinates in the trimmed playable grid.
Because multiple objects may occupy the same grid cell, overlap is represented by multiple object entries sharing the same \texttt{position}.
World objects are directional entities in the underlying simulator, and serialized entries expose this through a \texttt{direction} field when that orientation is available.
For example, a text block may be represented as \{\texttt{"type": "rule\_noun"}, \texttt{"word": "baba"}, \texttt{"position": [0,0]}\}, while a \texttt{world\_object} entry may additionally include a field such as \texttt{"direction": "facing right"}.

Actions are represented by a small discrete set of named commands, \texttt{idle}, \texttt{up}, \texttt{right}, \texttt{down}, and \texttt{left}.
Transitions are therefore recorded as structured triples consisting of a source-state JSON string, an action name, and a next-state JSON string.

\paragraph{Baba in Wonderland remapping.}

\begin{table}[t]
	\centering
	\small
\setlength{\tabcolsep}{2pt}
	\caption{Rule-property remapping used in \textit{Baba in Wonderland}. Simulator semantics remain unchanged.}
	\begin{tabular*}{\textwidth}{@{\hspace{4pt}\extracolsep{\fill}}llllllll@{\hspace{4pt}}}
		\toprule
		Original & Wonderland & Original & Wonderland & Original & Wonderland & Original & Wonderland \\
		\midrule
		\texttt{DEFEAT} & \texttt{WAKE} &
		\texttt{FLOAT} & \texttt{WRONG} &
		\texttt{HOT} & \texttt{GRIN} &
		\texttt{MELT} & \texttt{CURIOUS} \\
		\texttt{MOVE} & \texttt{DRINK} &
		\texttt{OPEN} & \texttt{MAD} &
		\texttt{PUSH} & \texttt{GROW} &
		\texttt{SHUT} & \texttt{LATE} \\
		\texttt{SINK} & \texttt{BEGIN} &
		\texttt{STOP} & \texttt{EAT} &
		\texttt{WIN} & \texttt{SHRINK} &
		\texttt{YOU} & \texttt{STRANGE} \\
		\bottomrule
	\end{tabular*}
	\label{tab:wonderland_remap}
\end{table}

The central intervention in \textit{Baba in Wonderland} is a remapping of the surface labels attached to \texttt{rule\_property} tokens.
This remapping changes only the words that appear in serialized states and transition prompts.
It does not alter the underlying simulator dynamics, the action space, or the mechanics by which rules are parsed and executed.
As a result, two environments that differ only by this remapping induce the same transition law even though they present different textual cues to the LLM.
The concrete remapping used in our experiments is listed in \autoref{tab:wonderland_remap}.

The purpose of this intervention is to break the direct alignment between pretrained lexical priors and the true behavioral meaning of rule-property tokens.
A model can no longer rely on the everyday meaning of an observed label to guess that it should behave like \texttt{YOU}, \texttt{WIN}, or \texttt{STOP}.
Instead, it must infer the role of each remapped token from transition evidence and from how that token participates in active rule sentences over time.
The agent is given no natural-language explanation of the remapping and no auxiliary dictionary that translates remapped labels back to their original meanings.

This design preserves the combinatorial and state-dependent structure of \textit{Baba Is You} while introducing prior misalignment directly into the observable rule vocabulary.
Success in \textit{Baba in Wonderland} therefore provides stronger evidence that the learned executable world model is grounded in interaction evidence rather than in semantically aligned textual shortcuts.

\clearpage 

\subsection{Prompt Templates}
\label{sec:appendix_prompt_templates}

\begin{figure*}[h!]
	\centering 
	\includegraphics[width=1\linewidth]{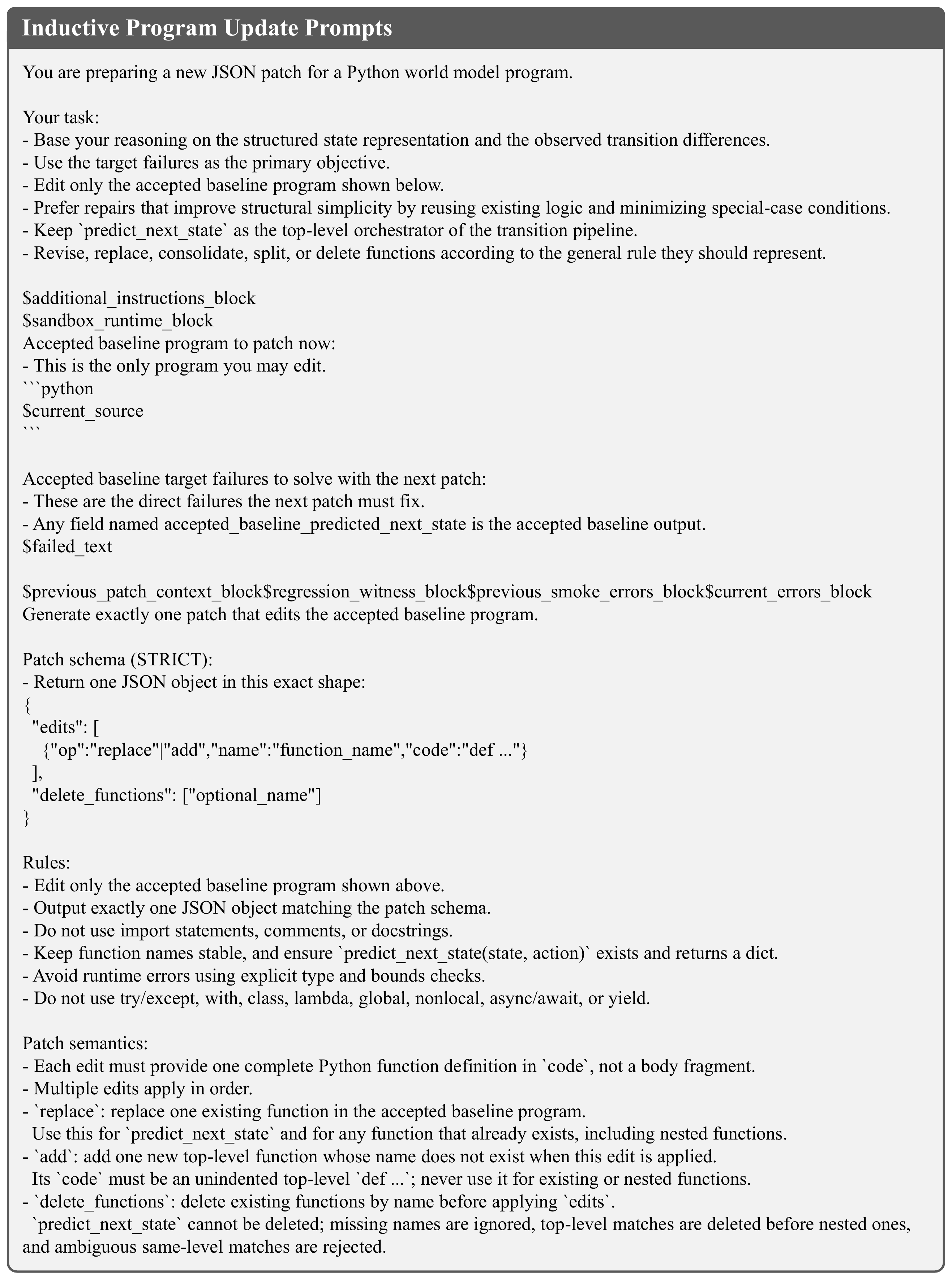}
	\caption{Inductive program update prompt template.}
	\label{fig:init_prompt_example}
\end{figure*}

\autoref{fig:init_prompt_example} shows the inductive program-update prompt template used by the Programmer.
The template provides the current program, the target explanatory failure, and broken examples selected from hypothesis classes.

\paragraph{Closed-loop procedure.}\leavevmode
Alice first finds a transition that its current program gets wrong. If a candidate fixes it but breaks previously explained examples, Alice uses those broken examples to guide the next update and organize exploration.
\begin{algorithm}[!htbp]

\caption{Alice: online executable world-model learning}
\label{alg:alice}
\begin{algorithmic}[1]
\Require Interaction budget $H$, initial program $P_0$, dataset $D$, classes $\mathcal C\leftarrow\emptyset$
\While{interaction and LLM-call budgets remain}
  \If{initial root classes have not formed}
    \State Select frontier candidates by BFS
  \Else
    \State Update the embedding on active classes when due
    \State Select candidates with highest $\lambda_h r_h+\lambda_{\mathcal C}r_{\mathcal C}$
  \EndIf
  \State Execute selected actions and add observed transitions to $D$
  \For{each transition $\tau$ not explained by $P_i$}
    \State $R\leftarrow\emptyset$
    \For{up to 15 update attempts, within the global call budget}
      \State $Q\leftarrow\mathrm{LLM}(P_i,\tau,R)$
      \If{$Q$ explains $\tau$ and preserves every transition in $E(P_i;D)$}
        \State Accept $Q$; update root classes and route new transitions; \textbf{break}
      \ElsIf{$Q$ explains $\tau$ but breaks previously explained transitions}
        \State Split partially affected classes into preserved and broken subsets
        \State $R\leftarrow$ one broken example from each of up to three affected classes
      \EndIf
    \EndFor
    \State Stop if the update fails; apply the remaining stopping rules in \appref{sec:appendix_experimental_details}
  \EndFor
\EndWhile
\end{algorithmic}
\end{algorithm}

\clearpage 

\subsection{Five-Run Analyses and Additional Evaluation}
\label{sec:appendix_revised_results}
\paragraph{Evaluation splits and reporting.}\leavevmode
All main comparisons and ablations report means and standard deviations over five independent runs. The independent unit is a complete run, including LLM updates, exploration, and termination, rather than an individual transition. Online learning collects evidence on the 32 regular levels; its primary evaluation uses regular-level solution transitions. Offline learning receives regular-level solution transitions and is evaluated on the eight extra levels. These protocols differ in both evidence source and evaluation set, so their accuracies do not establish that offline or online learning is intrinsically superior. The offline setting also removes the exploration burden by supplying solution transitions directly.

\paragraph{Class-coverage coefficient.}
We swept $\lambda_{\mathcal C}$ with $\lambda_h=1$ (\autoref{tab:appendix_lambda_sweep}). The five-run means are high over $\lambda_{\mathcal C}\in[0.2,0.5]$; we use $0.35$, which is near the center of this region and has the highest mean. This coefficient is fixed across the main online Alice experiments, including ALFWorld; the offline protocol has no Explorer. The sweep describes sensitivity within the evaluated configurations; it does not establish that every positive coefficient improves performance.

\begin{table}[!htbp]
\centering\small
\setlength{\tabcolsep}{2pt}
\caption{Class-coverage sensitivity on regular-level solution transitions in Wonderland, with $\lambda_h=1$. Five-run mean $\pm$ standard deviation.}
\label{tab:appendix_lambda_sweep}
\begin{tabular*}{\textwidth}{@{\hspace{4pt}\extracolsep{\fill}}lccc@{\hspace{4pt}}}
\toprule
Setting & All Acc. $\uparrow$ & Balanced Acc. $\uparrow$ & Calls \\
\midrule
\textbf{w/o} $r_{\mathcal C}$ & $0.929\pm0.050$ & $0.890\pm0.062$ & $87.0\pm20.5$ \\
$\lambda_{\mathcal C}=0.05$ & $0.928\pm0.050$ & $0.890\pm0.058$ & $82.0\pm18.9$ \\
$\lambda_{\mathcal C}=0.10$ & $0.953\pm0.045$ & $0.923\pm0.063$ & $96.2\pm8.5$ \\
$\lambda_{\mathcal C}=0.20$ & $0.953\pm0.045$ & $0.926\pm0.060$ & $100.0\pm0.0$ \\
$\lambda_{\mathcal C}=0.35$ (\textbf{Alice}) & $0.965\pm0.054$ & $0.951\pm0.068$ & $94.0\pm10.4$ \\
$\lambda_{\mathcal C}=0.50$ & $0.961\pm0.057$ & $0.937\pm0.091$ & $87.8\pm13.8$ \\
\bottomrule
\end{tabular*}
\end{table}

\paragraph{Extra-level evaluation of online programs.}
\autoref{tab:appendix_online_extra} evaluates the final programs from the same five online runs on extra-level solution transitions, without further program updates. All learning interactions come from regular levels. This evaluates generalization beyond the levels used for interaction. The main online comparison and the statistical tests use regular-level solution transitions for all methods.

\begin{table}[!htbp]
\centering\small
\setlength{\tabcolsep}{2pt}
\caption{Extra-level solution-transition accuracy of the online Alice runs. All rows use Wonderland unless stated otherwise. Calls are those of the regular-level learning runs. Five-run mean $\pm$ standard deviation.}
\label{tab:appendix_online_extra}
\begin{tabular*}{\textwidth}{@{\hspace{4pt}\extracolsep{\fill}}lccc@{\hspace{4pt}}}
\toprule
Setting & All Acc. $\uparrow$ & Balanced Acc. $\uparrow$ & Calls \\
\midrule
\textbf{Alice} ($\lambda_{\mathcal C}=0.35$) & $0.957\pm0.056$ & $0.945\pm0.067$ & $94.0\pm10.4$ \\
\textbf{Single-Class} & $0.877\pm0.065$ & $0.829\pm0.081$ & $84.8\pm20.9$ \\
\textbf{Root-Class} & $0.937\pm0.062$ & $0.911\pm0.084$ & $80.0\pm12.8$ \\
\textbf{w/o} $r_h$ & $0.924\pm0.046$ & $0.900\pm0.056$ & $83.0\pm13.5$ \\
\textbf{w/o} $r_{\mathcal C}$ & $0.902\pm0.071$ & $0.869\pm0.091$ & $87.0\pm20.5$ \\
\textbf{BFS} & $0.861\pm0.104$ & $0.811\pm0.136$ & $86.8\pm18.6$ \\
$\lambda_{\mathcal C}=0.05$ & $0.914\pm0.070$ & $0.886\pm0.084$ & $82.0\pm18.9$ \\
$\lambda_{\mathcal C}=0.10$ & $0.926\pm0.073$ & $0.898\pm0.097$ & $96.2\pm8.5$ \\
$\lambda_{\mathcal C}=0.20$ & $0.950\pm0.067$ & $0.930\pm0.088$ & $100.0\pm0.0$ \\
$\lambda_{\mathcal C}=0.50$ & $0.943\pm0.116$ & $0.922\pm0.153$ & $87.8\pm13.8$ \\
\textbf{Alice} (Default World) & $0.999\pm0.002$ & $0.999\pm0.002$ & $19.8\pm6.6$ \\
\bottomrule
\end{tabular*}
\end{table}

\paragraph{Statistical tests for the ablations.}
All tests use the regular-level evaluation and the run as the independent unit. For each Alice--variant comparison, we enumerate all $\binom{10}{5}=252$ assignments in an exact two-sided permutation test, applying Holm correction across the five comparisons separately for each metric. Exact Mann--Whitney tests give the same significance conclusions after correction. No individual comparison is significant after Holm correction at $\alpha=0.05$ (\autoref{tab:appendix_pairwise_tests}). The smallest attainable two-sided permutation $p$-value with five runs per group is $2/252\approx0.008$; the small run count limits resolution. Differences below are calculated before rounding the reported means.

\begin{table}[!htbp]
\centering\small
\setlength{\tabcolsep}{2pt}
\caption{Pairwise Alice--variant differences and exact two-sided permutation $p$-values. Holm correction is over five comparisons per metric.}
\label{tab:appendix_pairwise_tests}
\begin{tabular*}{\textwidth}{@{\hspace{4pt}\extracolsep{\fill}}lcccc@{\hspace{4pt}}}
\toprule
& \multicolumn{2}{c}{All Acc.} & \multicolumn{2}{c}{Balanced Acc.} \\
\cmidrule(lr){2-3}\cmidrule(lr){4-5}
Variant & $\Delta$ & $p$ (raw / Holm) & $\Delta$ & $p$ (raw / Holm) \\
\midrule
\textbf{Single-Class} & $+0.075$ & $.056/.278$ & $+0.122$ & $.024/.119$ \\
\textbf{Root-Class} & $+0.025$ & $.460/.919$ & $+0.044$ & $.357/.714$ \\
\textbf{w/o} $r_h$ & $+0.012$ & $.516/.919$ & $+0.026$ & $.492/.714$ \\
\textbf{w/o} $r_{\mathcal C}$ & $+0.037$ & $.230/.690$ & $+0.061$ & $.183/.549$ \\
\textbf{BFS} & $+0.064$ & $.071/.283$ & $+0.096$ & $.079/.317$ \\
\bottomrule
\end{tabular*}
\end{table}

We also test three orderings induced by progressively adding class structure or exploration signals. Each exact one-sided Jonckheere--Terpstra permutation trend test enumerates all $15!/(5!)^3=756{,}756$ assignments of the fifteen runs, with Holm correction across the three orderings separately for each metric. All three trends remain significant on both metrics (\autoref{tab:appendix_trend_tests}). This supports an overall monotonic trend, not a significant improvement for every adjacent pair or every individual component. In particular, $\text{w/o }r_h$ and $\text{w/o }r_{\mathcal C}$ form two separate paths from BFS to Alice; they are not ordered relative to one another.

\begin{table}[!htbp]
\centering\small
\setlength{\tabcolsep}{2pt}
\caption{Exact one-sided permutation trend tests; entries are raw / Holm-adjusted $p$-values.}
\label{tab:appendix_trend_tests}
\begin{tabular*}{\textwidth}{@{\hspace{4pt}\extracolsep{\fill}}lcc@{\hspace{4pt}}}
\toprule
Ordered hypothesis & All Acc. & Balanced Acc. \\
\midrule
\textbf{Single-Class} $<$ \textbf{Root-Class} $<$ \textbf{Alice} & $.006/.019$ & $.004/.012$ \\
\textbf{BFS} $<$ \textbf{w/o} $r_{\mathcal C}$ $<$ \textbf{Alice} & $.010/.019$ & $.028/.043$ \\
\textbf{BFS} $<$ \textbf{w/o} $r_h$ $<$ \textbf{Alice} & $.009/.019$ & $.021/.043$ \\
\bottomrule
\end{tabular*}
\end{table}

\paragraph{Same-call-cap WorldCoder diagnostic.}
Under the strict protocol, an online run terminates after exhausting the 15-call budget on a single failed update. To separate this stopping rule from the global call cap, the diagnostic retains WorldCoder's last accepted program after such a failure and continues collecting evidence and attempting later updates up to 100 calls. Any unfinished update at the cap is discarded. This changes the continuation rule; it matches the global cap, not the realized number of calls. \autoref{tab:appendix_worldcoder_equal_budget} reports five-run results. WorldCoder improves from $0.542$ to $0.819$ All Acc. and from $0.260$ to $0.658$ Balanced Acc., showing that early termination accounts for part of the strict-protocol gap. Alice still has a $0.293$ higher mean Balanced Acc. under the same cap.

\begin{table}[!htbp]
\centering\small
\setlength{\tabcolsep}{2pt}
\caption{Same-100-call-cap diagnostic on regular-level solution transitions in Wonderland. Five-run mean $\pm$ standard deviation.}
\label{tab:appendix_worldcoder_equal_budget}
\begin{tabular*}{\textwidth}{@{\hspace{4pt}\extracolsep{\fill}}lccc@{\hspace{4pt}}}
\toprule
Method & All Acc. $\uparrow$ & Balanced Acc. $\uparrow$ & Calls \\
\midrule
\textbf{Alice} & $0.965\pm0.054$ & $0.951\pm0.068$ & $94.0\pm10.4$ \\
\textbf{WorldCoder, continued} & $0.819\pm0.052$ & $0.658\pm0.131$ & $100.0\pm0.0$ \\
\bottomrule
\end{tabular*}
\end{table}

\clearpage
\subsection{Information-Theoretic View of Frontier Scoring}
\label{sec:appendix_frontier_score}

This section gives an information-theoretic view of the frontier score used by the Explorer.
The score is designed to make frontier coverage program-relative: the Explorer should cover frontier candidates in a way that is informative about the hypothesis classes induced by the accepted program-update sequence.
Let \(C_{\tau_x}\in\mathcal{C}\) denote the hypothesis class assigned to the induced transition \(\tau_x\) after executing frontier candidate \(x\).
This motivates the objective
\[
\max_{\pi}
H_\pi(x)+I_\pi(C_{\tau_x};x).
\]
The entropy term \(H_\pi(x)\) expands coverage over frontier candidates, but by itself it may only increase surface variation among state-action pairs.
The mutual-information term \(I_\pi(C_{\tau_x};x)\) makes this coverage class-informative: when \(x\) carries information about \(C_{\tau_x}\), increasing coverage over \(x\) is aligned with increasing coverage over hypothesis-class evidence.

This form follows two ideas from reward-free exploration and skill discovery.
APT motivates entropy-based coverage as a reward-free exploration objective~\citep{liu2021behavior}.
DIAYN and CIC motivate mutual information as a way to make a diversity variable recoverable from generated experience or learned representations~\citep{eysenbach2018diversity,laskin2022unsupervised}.
Alice changes the role of this variable: \(C_{\tau_x}\) is not a skill, option, or behavior variable to be executed, but a hypothesis class induced by accepted program updates.
The purpose is therefore not to learn a reusable skill repertoire, but to make frontier coverage relative to the current executable program.

Because \(x=(s,a)\) is a structured state-action object, we formulate the criterion in a learned representation space, with \(h=h_\theta(x)\).
The representation-level objective is
\[
\max_{\pi,\theta}
H_\pi(h)+I_\pi(C_{\tau_x};h).
\]
The role of \(h_\theta\) is to preserve hypothesis-class distinctions while providing a space in which coverage can be estimated.

Using the mutual-information identity,
\[
I_\pi(C_{\tau_x};h)
=
H_\pi(C_{\tau_x})-H_\pi(C_{\tau_x}\mid h),
\]
we obtain
\[
J_{\mathrm{explore}}
=
H_\pi(h)-H_\pi(C_{\tau_x}\mid h)+H_\pi(C_{\tau_x}).
\]
Maximizing this objective imposes three pressures.
The first term, \(H_\pi(h)\), encourages coverage in the learned embedding space.
The second term, \(-H_\pi(C_{\tau_x}\mid h)\), makes the embedding class-informative.
The final term, \(H_\pi(C_{\tau_x})\), encourages coverage across hypothesis classes, preventing the policy from satisfying the objective by expanding variation inside only one or a few classes.

The frontier score in \autoref{sec:update_guided_exploration} is a computable approximation of this criterion.
The embedding-novelty score \(r_h(x)\) approximates the \(H_\pi(h)\) term, the active-class contrastive objective supports the \(-H_\pi(C_{\tau_x}\mid h)\) term, and the expected class-coverage score \(r_{\mathcal{C}}(x)\) approximates the \(H_\pi(C_{\tau_x})\) term.
Thus,
\[
S_{\mathrm{frontier}}(x)
=
\lambda_h r_h(x)
+
\lambda_{\mathcal{C}} r_{\mathcal{C}}(x)
\]
can be viewed as a practical frontier-scoring approximation to the entropy--mutual-information objective.

\clearpage

\subsection{Additional Analyses}
\label{sec:appendix_additional_experiments}
This subsection complements the main experiments with analyses of learning behavior, class alignment, embedding structure, transition coverage, update-evidence hyperparameters, and LLM backbones. The learning curves average the five runs used in the main ablations; the other analyses use a single run per configuration.

\begin{figure*}[!htbp]
\centering
	\begin{minipage}[c]{0.49\textwidth}
		\centering
		\includegraphics[width=\linewidth]{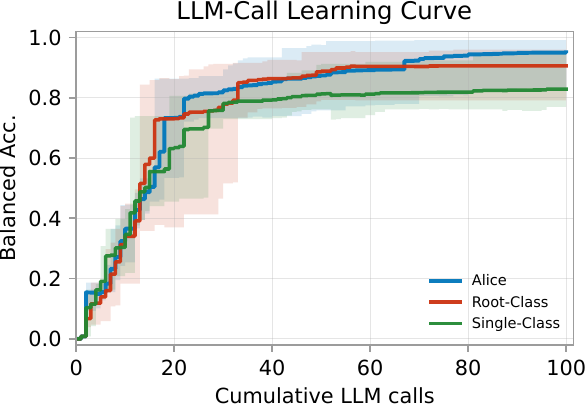}
	\end{minipage}\hfill
	\begin{minipage}[c]{0.49\textwidth}
		\centering
		\includegraphics[width=\linewidth]{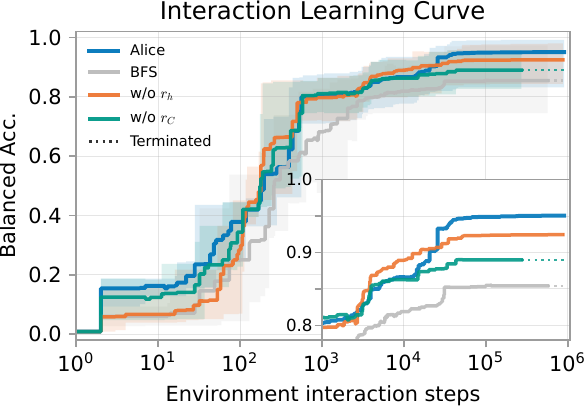}
	\end{minipage}
	\caption{Learning curves averaged over five runs; shading spans the minimum and maximum over runs, and a stopped run keeps its final accuracy. Left: hypothesis-class variants over cumulative LLM calls. Right: exploration variants over interaction steps; a dotted line starts once all five runs of a method have stopped.}
	\label{fig:wonderland_ablation_curves}
\end{figure*}

\begin{table*}[t]
	\centering
	\small
\setlength{\tabcolsep}{2pt}
	\caption{Hypothesis-class alignment for Alice. Purity is measured from heuristic transition classes \(\mathcal{C}_H\) to Alice's learned classes \(\mathcal{C}\) on human-play solution transitions.}
	\begin{tabular*}{\textwidth}{@{\hspace{4pt}\extracolsep{\fill}}llccc@{\hspace{4pt}}}
		\toprule
		Setting
		& Evaluation Set
		& Purity $\uparrow$
		& \(\#\mathcal{C}_H\)
		& \(\#\mathcal{C}\) \\
		\midrule
		Default World & Regular levels & 0.978 & 595 & 23 \\
		Default World & Extra levels   & 0.955 & 293 & 14 \\
		Baba in Wonderland & Regular levels & 0.793 & 595 & 66 \\
		Baba in Wonderland & Extra levels   & 0.817 & 293 & 36 \\
		\bottomrule
	\end{tabular*}
	\label{tab:appendix_class_alignment}
\end{table*}

\paragraph{Learning curves.} \autoref{fig:wonderland_ablation_curves} (left) shows how the hypothesis-class variants fall behind. Two of the five Single-Class runs exhaust the 15-call budget on a single update and stop at 59 and 65 calls. Without classes, the three broken examples in the update context are drawn at random from all broken transitions, so they can repeat one preservation constraint while missing the others that a candidate breaks. The other three Single-Class runs accept 24--28 additional updates per run after call 40, compared with 17--30 per run for Alice, but improve Balanced Acc. by only $0.057$ on average, compared with $0.097$ for Alice. With a single class, the class-coverage term is zero, so the Explorer cannot direct evidence toward weakly supported parts of the program. Root-Class tracks Alice up to about 66 calls, and their mean curves cross several times. Early root classes still hold few distinct constraints, so one randomly drawn broken example per root class is often as informative as one per refined class. As accepted updates add transitions to these root classes, each comes to mix several constraints, and a random draw more often misses the one that a candidate breaks. Three of the five Root-Class runs then exhaust the per-update budget, against none for Alice, and Alice ends above Root-Class for every seed.

\autoref{fig:wonderland_ablation_curves} (right) shows a similar late separation for exploration. All frontier-scoring explorers reach about $0.80$ Balanced Acc. by $10^3$ steps, against $0.68$ for BFS. The two score ablations stay close to Alice up to about $2.5\times10^4$ steps. Between $10^4$ and $10^5$ steps, Alice gains $0.083$ in Balanced Acc., compared with $0.027$--$0.036$ for the ablations. The score-ablation runs also stop earlier: the median run ends after $2.6\times10^4$ steps without \(r_h\) and $3.4\times10^4$ steps without \(r_C\), against $5.2\times10^5$ for Alice. Two or three of the five runs in each ablation, including BFS, exhaust the per-update budget, against none for Alice. Once common transitions are explained, the remaining errors involve rare dynamics, and reaching them appears to require both novelty and coverage of weakly supported classes.

\paragraph{Hypothesis-class alignment.}
We further test whether Alice's learned hypothesis classes remain tied to transition structure when semantic shortcuts are removed. For this diagnostic, we evaluate Alice's classes \(\mathcal{C}\) on the solution transition dataset and compare them with an evaluation-only heuristic partition \(\mathcal{C}_H\), which groups transitions by action and canonical state-difference signature and is never used during learning. We report micro-averaged one-way purity from \(\mathcal{C}_H\) to \(\mathcal{C}\): for each heuristic class, we assign its evaluated transitions to their majority Alice class and aggregate the majority counts across heuristic classes. The reported \(\#\mathcal{C}\) is therefore the number of Alice classes represented among the evaluated solution transitions, not the total number of classes discovered in the broader exploration archive. As shown in \autoref{tab:appendix_class_alignment}, purity remains substantial in both \textit{Default World} and \textit{Baba in Wonderland}, despite a drop under the misleading Wonderland labels. This indicates that transitions with the same behavioral effect remain largely grouped under Alice's learned partition, even when the surface rule-property labels no longer provide reliable semantic cues.

\begin{figure*}[t]
	\centering
	\begin{minipage}[c]{0.246\textwidth}
		\centering
		\includegraphics[width=\linewidth]{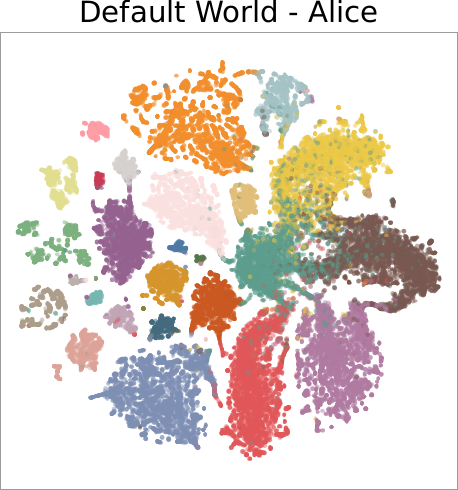}
	\end{minipage}\hfill
	\begin{minipage}[c]{0.246\textwidth}
		\centering
		\includegraphics[width=\linewidth]{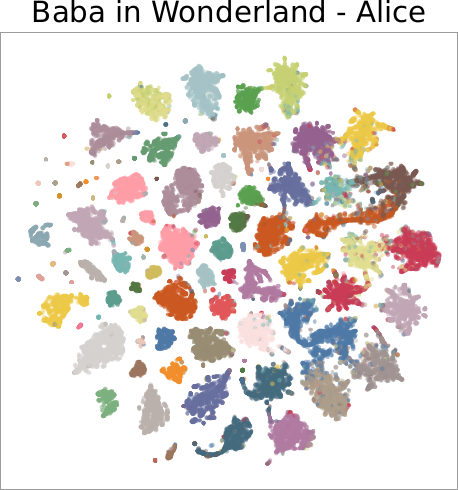}
	\end{minipage}\hfill
	\begin{minipage}[c]{0.246\textwidth}
		\centering
		\includegraphics[width=\linewidth]{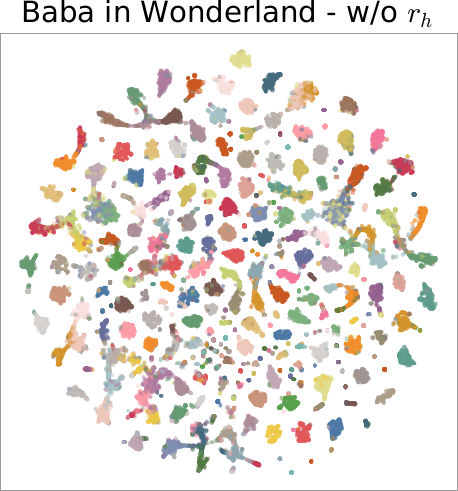}
	\end{minipage}\hfill
	\begin{minipage}[c]{0.246\textwidth}
		\centering
		\includegraphics[width=\linewidth]{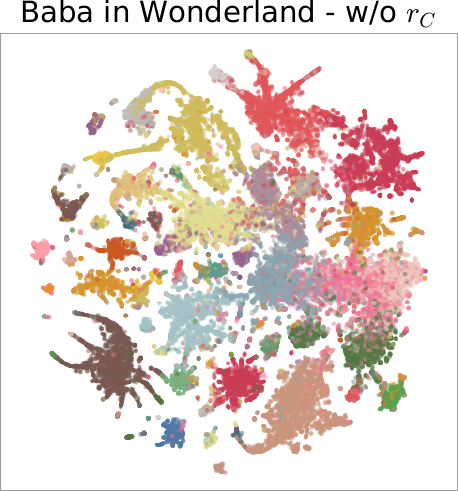}
	\end{minipage}
	\caption{
	t-SNE visualization of the Explorer's learned state-action embedding space.
	Points are explained transitions projected from the \(384\)-dimensional embedding space, and colors denote assigned hypothesis classes.
	From left to right: Alice in \textit{Default World}, Alice in \textit{Baba in Wonderland}, \textbf{w/o} \(r_h\) in \textit{Baba in Wonderland}, and \textbf{w/o} \(r_C\) in \textit{Baba in Wonderland}.
	}
	\label{fig:appendix_tsne}
\end{figure*}

\paragraph{Embedding space t-SNE visualization.}
\label{sec:appendix_embedding}
\leavevmode\autoref{fig:appendix_tsne} provides a qualitative view of the learned state-action embeddings. Each panel samples 50,000 explained transitions, projects their 384-dimensional embeddings with t-SNE, and colors points by assigned class. Alice forms class-structured embeddings in both label settings. Removing either exploration term changes the distribution of visited classes: without \(r_h\), the view contains smaller fragmented clusters; without \(r_C\), samples concentrate in a few larger classes. These views complement the exploration ablations by illustrating the organization of collected evidence; distances in the projection do not directly measure state-space coverage.

\begin{table*}[t]
	\centering
	\small
\setlength{\tabcolsep}{2pt}
	\caption{Evaluation on the held-out extra-level coverage transition dataset. All Acc. is measured on the full coverage transition dataset; Balanced Acc. is measured on the corresponding class-reduced coverage transition set.}
	\begin{tabular*}{\textwidth}{@{\hspace{4pt}\extracolsep{\fill}}lcccc@{\hspace{4pt}}}
		\toprule
		& \multicolumn{2}{c}{Default World}
		& \multicolumn{2}{c}{Baba in Wonderland} \\
		\cmidrule(lr){2-3} \cmidrule(lr){4-5}
		Method
		& All Acc. $\uparrow$
		& Balanced Acc. $\uparrow$
		& All Acc. $\uparrow$
		& Balanced Acc. $\uparrow$ \\
		\midrule
		\textbf{WorldCoder} & 0.984 & 0.894 & 0.749 & 0.029 \\
		\textbf{Alice}      & 0.999 & 0.996 & 0.997 & 0.995 \\
		\bottomrule
	\end{tabular*}
	\label{tab:appendix_broad_coverage}
\end{table*}

\paragraph{Coverage transition dataset evaluation.}
The main experiments evaluate on human-play solution transitions, which concentrate on behavior encountered along successful solution trajectories.
As a complementary stress test, we evaluate the online runs trained on regular levels on the coverage transition dataset collected by BFS over each held-out extra level.
All Acc. uses the full coverage transition dataset, while Balanced Acc. uses the corresponding class-reduced coverage transition set to discount repeated local effects.
\autoref{tab:appendix_broad_coverage} shows that WorldCoder remains much more reliable in \textit{Default World} than in \textit{Baba in Wonderland}; the gap is especially large under Balanced Acc., indicating that lexical shortcuts can preserve common coverage transitions while failing to recover diverse transition effects under prior misalignment.
By contrast, Alice maintains high accuracy on the coverage transition dataset in \textit{Baba in Wonderland}, showing that its learned executable model transfers beyond human-play solution trajectories.

\begin{table*}[t]
	\centering
	\small
\setlength{\tabcolsep}{2pt}
	\caption{Offline ablation of Alice's program-update evidence hyperparameters in \textit{Baba in Wonderland}. The default Alice setting is \(n=3,m=1\).}
	\begin{tabular*}{\textwidth}{@{\hspace{4pt}\extracolsep{\fill}}lccc@{\hspace{4pt}}}
		\toprule
		Evidence Setting
		& All Acc. $\uparrow$
		& Balanced Acc. $\uparrow$
		& Calls \\
		\midrule
			\(n=1,m=1\)         & 0.722 & 0.621 & 51.0 \\
			\(n=1,m=3\)         & 0.857 & 0.809 & 100.0 \\
			\(n=3,m=1\) (\textbf{Alice}) & 0.956 & 0.932 & 100.0 \\
			\(n=3,m=3\)         & 0.964 & 0.911 & 100.0 \\
		\bottomrule
	\end{tabular*}
	\label{tab:appendix_update_evidence_hparams}
\end{table*}

\paragraph{Program-update evidence hyperparameters.}
Alice constructs the LLM update context by selecting up to \(n\) affected hypothesis classes and sampling up to \(m\) broken examples from each selected class.
The main experiments use \(n=3\) and \(m=1\).
We ablate these values in the offline fixed-dataset setting, where all methods train on fixed regular-level human-play transitions and are evaluated on held-out extra-level human-play transitions under a \(100\)-call cap.
\autoref{tab:appendix_update_evidence_hparams} shows that performance is sensitive to how selected broken examples are packaged.
The default \(n=3,m=1\) setting gives the strongest Balanced Acc., while \(n=3,m=3\) slightly improves All Acc. at the cost of lower Balanced Acc.
The \(n=1\) variants are substantially weaker, indicating that too little class diversity in the program update context can miss rare preservation constraints.
We therefore treat these results as a sensitivity analysis: class-selected broken examples are useful, but larger update contexts are not automatically better under a fixed LLM-call budget.

\begin{table*}[t]
	\centering
	\small
\setlength{\tabcolsep}{2pt}
	\caption{Alice with different LLM backbones in \textit{Baba in Wonderland}. Results are online regular-level human-play transition accuracy under the same exploration and update protocol.}
	\begin{tabular*}{\textwidth}{@{\hspace{4pt}\extracolsep{\fill}}lccc@{\hspace{4pt}}}
		\toprule
		Backbone
		& All Acc. $\uparrow$
		& Balanced Acc. $\uparrow$
		& Calls \\
		\midrule
		\textbf{GPT-5.4-mini}     & 0.865 & 0.713 & 41.0 \\
		\textbf{Gemini 3.0 Flash} & 0.958 & 0.903 & 56.0 \\
		\textbf{GPT-5.4}          & 0.982 & 0.973 & 100.0 \\
		\textbf{Gemini 3.1 Pro}   & 0.993 & 0.992 & 65.0 \\
		\bottomrule
	\end{tabular*}
	\label{tab:appendix_backbone_variation}
\end{table*}

\paragraph{LLM backbone variation.}
We evaluate Alice with different LLM backbones while keeping the prompting format, program-update loop, exploration strategy, and evaluation protocol fixed.
\autoref{tab:appendix_backbone_variation} reports online performance on regular-level human-play transitions in \textit{Baba in Wonderland}.
The runs show different update dynamics across backbones.
In our qualitative inspection, Gemini 3.1 Pro quickly produced a broad program that achieved high accuracy, but later had difficulty revising that program for a few special cases and terminated before using the full call budget.
GPT-5.4 instead tended to make more incremental program revisions and continued until the full \(100\)-call budget was exhausted.
The smaller-model runs terminated earlier and reached lower accuracy, especially under Balanced Acc.
These results suggest that LLM backbones can differ not only in final accuracy, but also in how they revise executable programs; Alice's class-structured update and exploration loop nevertheless remains usable across the tested backbones.

\subsection{Qualitative Examples}
\label{sec:appendix_qualitative_examples}

For readability, the visualizations in this subsection render states with the original \textit{Baba Is You} sprites, although the evaluated transitions and learned programs come from \textit{Baba in Wonderland}; only the textual rule-property labels are remapped in the agent's observations.

\begin{figure*}[t]
	\centering
	\includegraphics[width=1\linewidth]{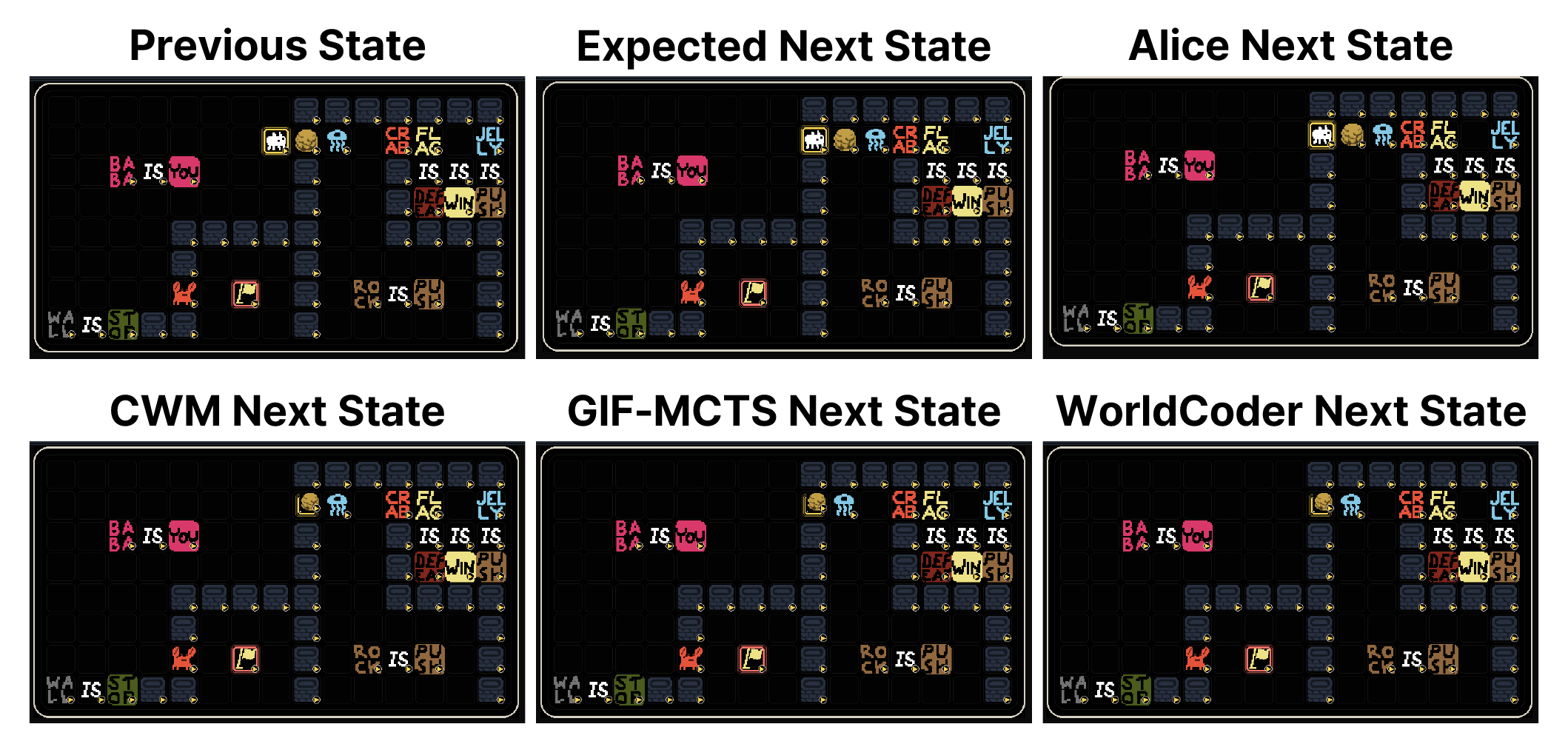}
	\caption{Success Case 1 from \textit{Baba in Wonderland}: a push chain under misleading rule-property labels. Each method panel shows the previous state, expected next state, and the method's predicted next state.}
	\label{fig:wonderland_success_case_study_01}
\end{figure*}

\begin{figure*}[t]
	\centering
	\includegraphics[width=1\linewidth]{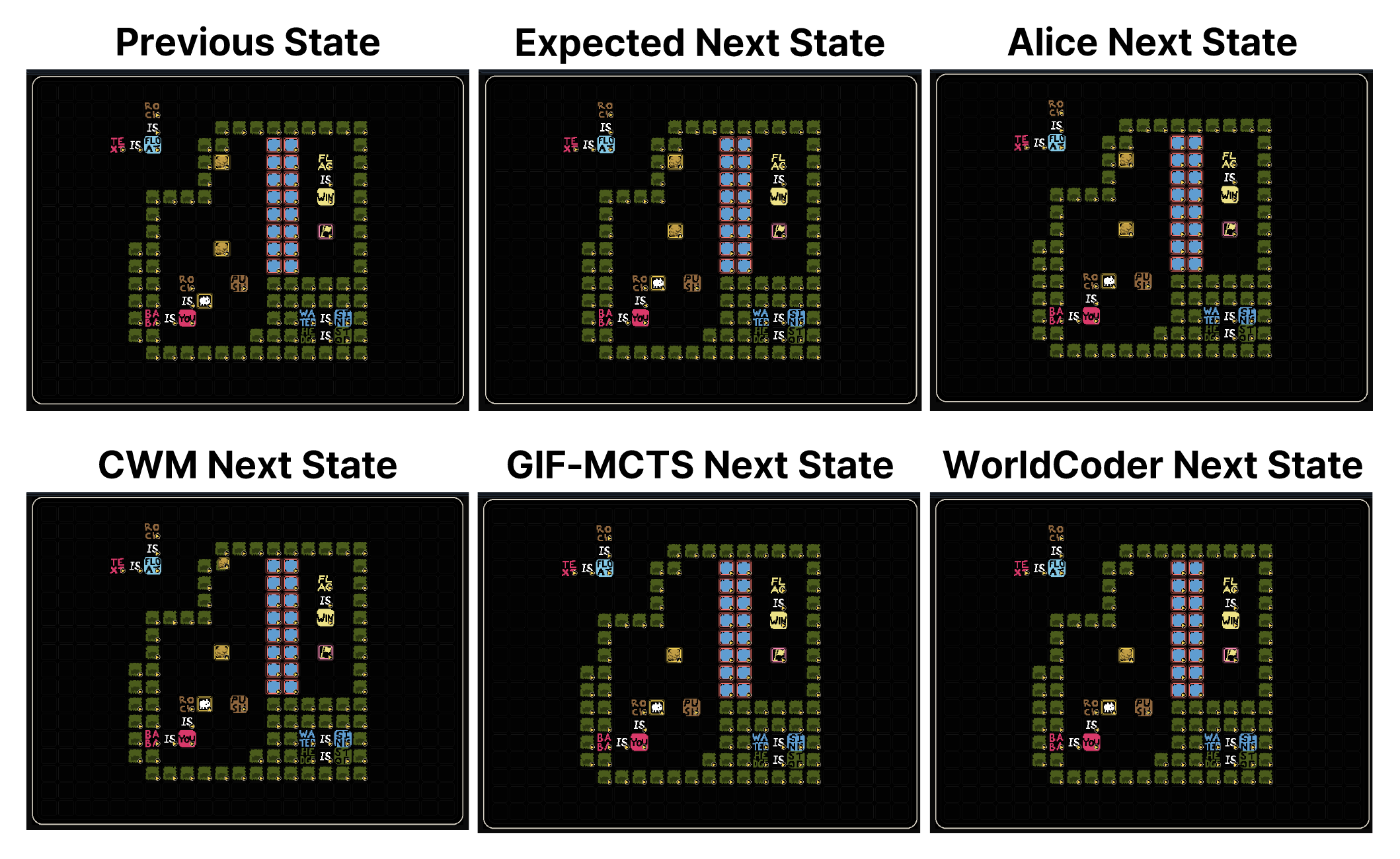}
	\caption{Success Case 2 from \textit{Baba in Wonderland}: simultaneous movement of multiple \texttt{YOU} objects with one controlled object blocked by a stop object. Each method panel shows the previous state, expected next state, and the method's predicted next state.}
	\label{fig:wonderland_success_case_study_02}
\end{figure*}

\paragraph{Success cases.}
\autoref{fig:wonderland_success_case_study_01} shows an evaluation transition on which Alice predicts the exact next state while all three baselines fail.
In the underlying game semantics, \texttt{BABA IS YOU}, \texttt{ROCK IS PUSH}, and \texttt{JELLY IS PUSH} are active; in \textit{Baba in Wonderland}, these appear as \texttt{BABA IS STRANGE}, \texttt{ROCK IS GROW}, and \texttt{JELLY IS GROW}.
When \texttt{BABA} moves right, the correct transition is a two-object push chain: \texttt{JELLY} moves into the empty cell, \texttt{ROCK} moves into the \texttt{JELLY}'s previous cell, and \texttt{BABA} moves into the \texttt{ROCK}'s previous cell.
Alice's learned program succeeds because it infers that the surface label \texttt{GROW} plays the role of \texttt{PUSH} and applies a recursive chain movement routine before moving the controlled object.
The selected WorldCoder program moves controlled objects but does not model this push-chain effect, while the GIF-MCTS and CWM programs contain generic push logic but do not identify the Wonderland \texttt{GROW} property as \texttt{PUSH} in this configuration.
All three baselines therefore leave the \texttt{ROCK}--\texttt{JELLY} chain behind, whereas Alice recovers the full transition from evidence despite the misleading property label.

\autoref{fig:wonderland_success_case_study_02} shows a different kind of success case involving simultaneous movement and blocking.
Here \texttt{BABA IS YOU} and \texttt{ROCK IS YOU} make both \texttt{BABA} and \texttt{ROCK} controlled objects, while \texttt{HEDGE IS STOP} makes \texttt{HEDGE} blocking objects; in \textit{Baba in Wonderland}, these rules appear as \texttt{BABA IS STRANGE}, \texttt{ROCK IS STRANGE}, and \texttt{HEDGE IS EAT}.
After the upward action, the lower controlled objects move up, but the upper \texttt{ROCK} is immediately below a \texttt{HEDGE} and should therefore remain in place while updating its facing direction.
Alice's learned program handles this mixed outcome: it moves each controlled object in the correct order, checks the blocking constraint for the \texttt{ROCK} adjacent to the \texttt{HEDGE}, and preserves the blocked \texttt{ROCK} rather than deleting it or moving it into the \texttt{HEDGE} cell.
The baselines miss this interaction in different ways: WorldCoder lacks the corresponding stop-blocking rule, GIF-MCTS omits the blocked \texttt{ROCK}, and CWM moves the \texttt{ROCK} into the blocked cell.
This case illustrates why balanced transition accuracy is diagnostic: the difficult part is not moving a controlled object in isolation, but preserving the exact simulator semantics when several controlled objects interact with blocking structure at the same time.

\begin{figure*}[t]
	\centering
	\includegraphics[width=1\linewidth]{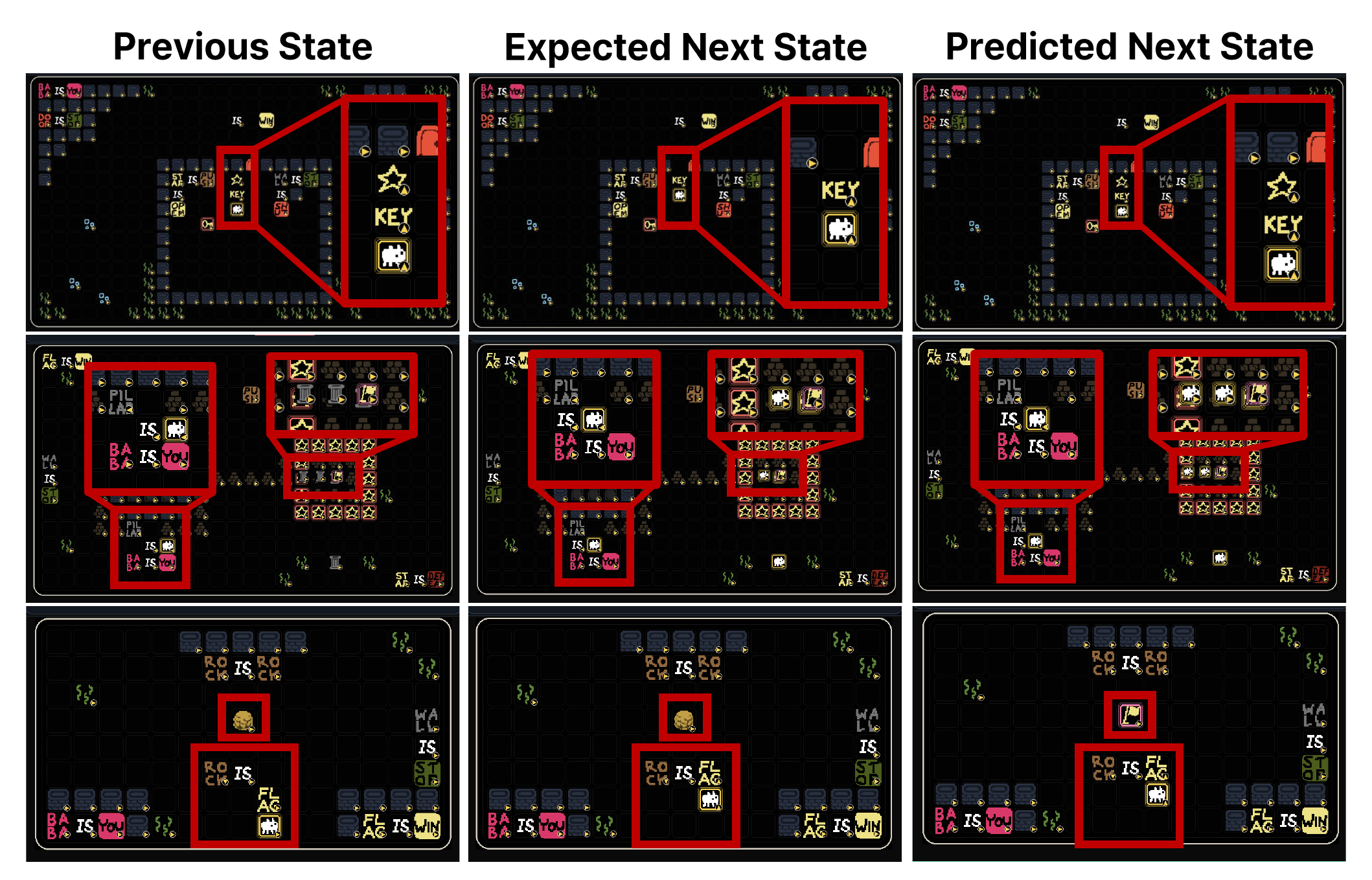}
	\caption{Failure case studies from \textit{Baba in Wonderland}.}
	\label{fig:wonderland_failure_case_studies}
\end{figure*}

\paragraph{Failure cases.}
A representative failure case is shown in the top row of \autoref{fig:wonderland_failure_case_studies}. 
In this configuration, \texttt{WALL IS SHUT}, \texttt{STAR IS OPEN}, and \texttt{STAR IS PUSH} are active; in \textit{Baba in Wonderland}, these appear as \texttt{WALL IS LATE}, \texttt{STAR IS MAD}, and \texttt{STAR IS GROW}. 
When \texttt{BABA} pushes the \texttt{KEY} text block, the pushed block in turn displaces the \texttt{STAR} into the \texttt{WALL}. 
Under the simulator dynamics, the resulting \texttt{OPEN}--\texttt{SHUT} interaction should remove both the \texttt{STAR} and the \texttt{WALL}, even though the \texttt{STAR} was moved indirectly rather than directly by a \texttt{YOU} object. 
Alice's learned program fails to predict this cancellation and leaves the objects in place. 
The error comes from an over-simplified collision routine in the induced program: because many important interactions in the data are triggered by agent-controlled movement, the learned code checks object--object interaction effects primarily for motion initiated by \texttt{YOU}-controlled objects. 
In this case, however, the relevant overlap is created by a \texttt{STAR} that is displaced by another pushed object, not by \texttt{BABA} itself. 
As a result, the learned program misses the \texttt{OPEN}--\texttt{SHUT} resolution that the simulator applies, illustrating how Alice can fit a compact but overly narrow causal account of collision dynamics.

A second failure case is shown in the middle row of \autoref{fig:wonderland_failure_case_studies}. 
Here a noun-to-noun transformation rule of the form \texttt{PILLAR IS BABA} becomes active, and the learned program correctly converts the \texttt{PILLAR} objects into \texttt{BABA}. 
However, one of the transformed \texttt{PILLAR} objects occupies a cell already overlapping a \texttt{STAR} object with the \texttt{DEFEAT} property. 
Under the simulator dynamics, the newly created \texttt{BABA} should then be removed immediately, because \texttt{BABA} is a \texttt{YOU} object and thus dies on overlap with \texttt{DEFEAT}. 
Alice's learned program performs the noun-to-noun conversion but does not resolve the resulting world-object interaction afterward, so the transformed \texttt{BABA} incorrectly remains in the predicted next state. 
This error reflects a highly specific blind spot in the induced program: it models object transformation and object--object collision as partially separate cases, but does not apply collision resolution after a fresh noun-to-noun conversion. 
Because such transitions are rare in the learning data, the resulting program never acquires the more complete rule that conversion can immediately trigger post-transform interaction effects.

A third failure case is shown in the bottom row of \autoref{fig:wonderland_failure_case_studies}. 
In this configuration, \texttt{ROCK IS ROCK} is active when a competing noun-to-noun rule \texttt{ROCK IS FLAG} is formed. 
Under the simulator dynamics, the self-identity rule takes precedence: once \texttt{A IS A} is active, it blocks other noun-to-noun rewrites for that object, so the existing \texttt{ROCK} objects should remain \texttt{ROCK} rather than transforming into \texttt{FLAG}. 
Alice's learned program instead applies \texttt{ROCK IS FLAG} and converts the \texttt{ROCK} objects into \texttt{FLAG}, producing an incorrect next state. 
This failure reflects a missing rule-ordering constraint in the induced program. 
From observed transitions alone, it is difficult to infer that self-identity rules are not merely redundant transformations but priority-setting constraints that suppress competing noun conversions. 
Because this precedence relation appears only in relatively specific configurations, the learned program captures the surface conversion pattern but not the underlying execution order that makes \texttt{ROCK IS ROCK} dominate \texttt{ROCK IS FLAG}.

\begin{figure*}[t]
	\centering
	\includegraphics[width=\textwidth]{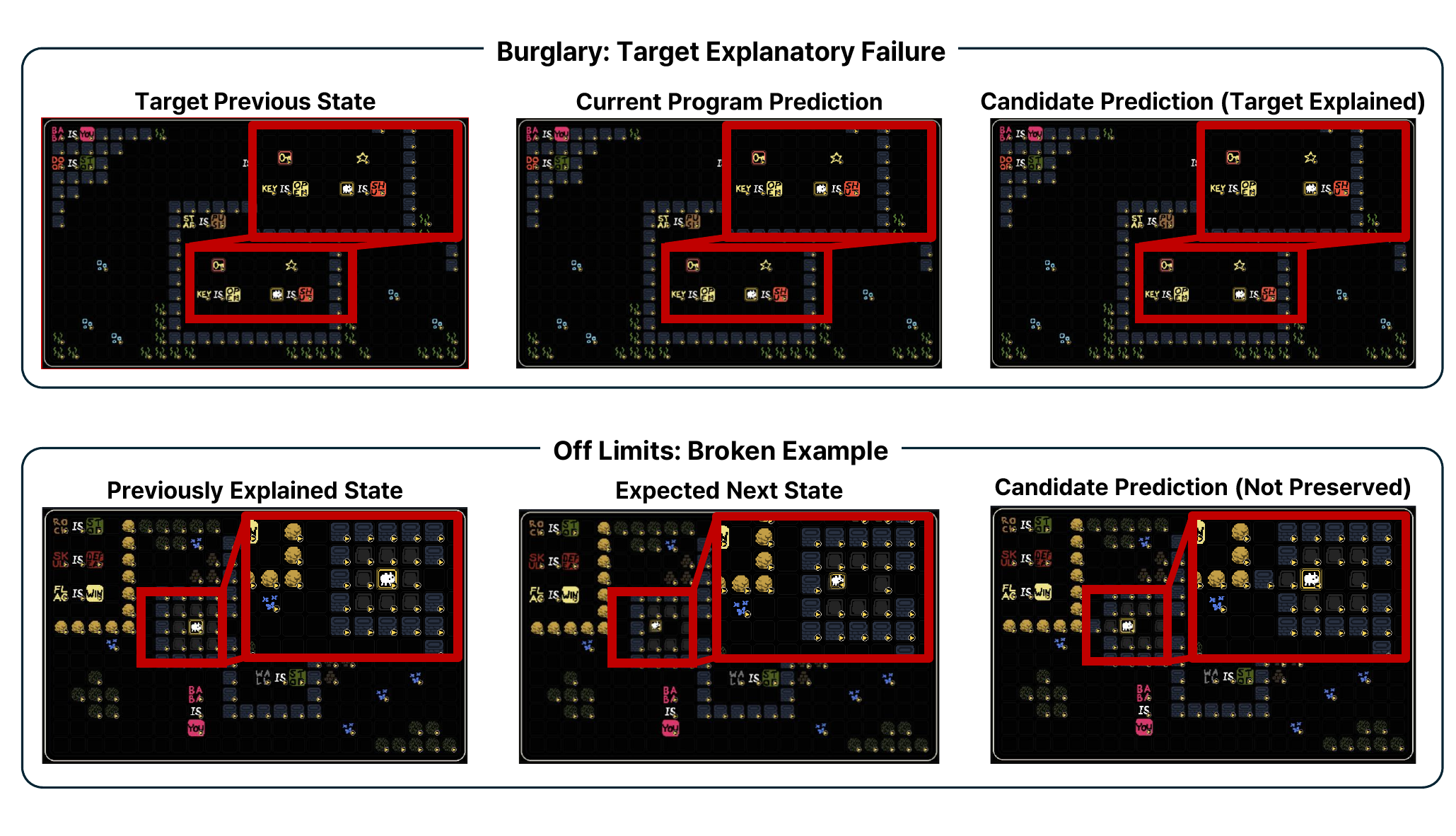}
	\caption{Actual rejected candidate update from Alice in \textit{Baba in Wonderland}. Top: the candidate explains the target transition from \textit{Burglary}, where \texttt{BABA} pushes adjacent rule text. Bottom: the same candidate fails to preserve a previously explained transition from \textit{Off Limits}, exposing a broken example for the next update prompt.}
	\label{fig:appendix_update_trace}
\end{figure*}

\paragraph{Example update trace.}
\autoref{fig:appendix_update_trace} makes the class-refinement mechanism concrete through one actual update trace from Alice in \textit{Baba in Wonderland}.
Alice reaches a transition in the level \textit{Burglary} where \texttt{BABA IS STRANGE} (\texttt{YOU}) is active and the action is \texttt{right}.
The correct next state moves \texttt{BABA} one cell to the right, but it also moves the adjacent rule-text chain \texttt{IS LATE} one cell to the right.
Here \texttt{LATE} is the Wonderland surface label for \texttt{SHUT}.
The current program explains ordinary controlled movement, but it does not yet model this case where a controlled object pushes rule text.
The transition therefore becomes an explanatory failure.

The first candidate update fixes this target: it moves both \texttt{BABA} and the adjacent text chain.
However, the executable acceptance check also tests whether the candidate preserves transitions that the previous program already explained.
The candidate fails this check.
In a previously explained transition from the level \textit{Off Limits}, \texttt{BABA IS STRANGE} (\texttt{YOU}) is active and the action is \texttt{left}.
The correct next state moves only \texttt{BABA} one cell left.
The candidate instead overgeneralizes the new push rule and also moves nearby non-controlled objects, including a \texttt{TILE} and a \texttt{WALL}.
Thus, the candidate has learned a useful part of the target behavior, but it has confused ``pushing adjacent rule text'' with a broader rule that moves unrelated neighboring objects.

Alice uses this rejection to refine its hypothesis classes.
The affected class is split into transitions that the rejected candidate still preserves and transitions that it breaks.
The preserved subset contains cases compatible with the candidate's new behavior; the lost subset contains broken examples showing where that behavior overgeneralizes.
Alice samples a representative lost transition, such as the \textit{Off Limits} case above, and includes it in the next update prompt as a selected broken example.
The next candidate is therefore asked to explain the original \textit{Burglary} failure while also preserving the \textit{Off Limits} broken example.
That retry is accepted with no preservation losses.
The trace in \autoref{fig:appendix_update_trace} shows why rejected updates are informative in Alice: they reveal which transitions were incorrectly conflated, split the corresponding hypothesis class, and determine which broken examples the next LLM update sees.

\subsection{Experimental Details}
\label{sec:appendix_experimental_details}

\begin{table*}[t]
	\centering
	\small
\setlength{\tabcolsep}{2pt}
	\caption{Original \textit{Baba Is You} level display names used in the regular and extra level sets.}
	\begin{tabularx}{\textwidth}{@{\hspace{4pt}}l>{\raggedright\arraybackslash}X>{\raggedright\arraybackslash}X@{\hspace{4pt}}}
		\toprule
		Stage & Regular Levels & Extra Levels \\
		\midrule
		1 &
		\textit{Baba Is You}; \textit{Where Do I Go?}; \textit{Now What Is This?}; \textit{Out Of Reach}; \textit{Still Out Of Reach}; \textit{Volcano}; \textit{Off Limits}; \textit{Grass Yard} &
		-- \\
		2 &
		\textit{Icy Waters}; \textit{Turns}; \textit{Affection}; \textit{Pillar Yard}; \textit{Brick Wall}; \textit{Lock}; \textit{Novice Locksmith}; \textit{Locked In}; \textit{Changeless}; \textit{Two Doors}; \textit{Jelly Throne}; \textit{Crab Storage}; \textit{Burglary} &
		\textit{Submerged Ruins}; \textit{Sunken Temple} \\
		3 &
		\textit{Float}; \textit{Warm River}; \textit{Bridge Building}; \textit{Bridge Building?}; \textit{Victory Spring}; \textit{Assembly Team}; \textit{Catch The Thief!}; \textit{Tiny Pond}; \textit{Research Facility}; \textit{Wireless Connection}; \textit{Prison} &
		\textit{Boiling River}; \textit{\ldots Bridges?}; \textit{Tiny Isle}; \textit{Dim Signal}; \textit{Dungeon}; \textit{Evaporating River} \\
		\bottomrule
	\end{tabularx}
	\label{tab:appendix_level_names}
\end{table*}

\paragraph{Regular and extra levels.}
We use original \textit{Baba Is You} levels drawn from Stages 1, 2, and 3 of the base game.
Following the level taxonomy of the game, we distinguish \textit{regular levels} from \textit{extra levels}; this partition is defined over levels, not by splitting trajectories from the same level.
The regular set contains \(32\) levels: all \(8\) Stage-1 levels, \(13\) regular Stage-2 levels, and \(11\) regular Stage-3 levels.
The extra set contains \(8\) levels: \(2\) Stage-2 extra levels and \(6\) Stage-3 extra levels.
The level display names in each set are listed in \autoref{tab:appendix_level_names}.
Online experiments collect interaction data on regular levels and evaluate on human-play solution transitions from those same levels, whereas offline fixed-dataset experiments train on regular-level human-play transitions and evaluate on held-out extra-level human-play transitions.

\begin{table}[t]
	\centering
	\small
\setlength{\tabcolsep}{2pt}
	\caption{Archived solution and coverage transition dataset statistics after canonical transition deduplication.}
	\begin{tabular*}{\textwidth}{@{\hspace{4pt}\extracolsep{\fill}}llccc@{\hspace{4pt}}}
		\toprule
		Split & Dataset & \# Levels & \# Transitions & \# States \\
		\midrule
		Regular Levels & Solution transition dataset & 32 & 2{,}476 & 2{,}490 \\
		Regular Levels & Coverage transition dataset & 32 & 3{,}200{,}000 & 1{,}254{,}186 \\
		Extra Levels  & Solution transition dataset & 8 & 1{,}126 & 1{,}126 \\
		Extra Levels  & Coverage transition dataset & 8 & 800{,}000 & 306{,}488 \\
		\bottomrule
	\end{tabular*}
	\label{tab:appendix_dataset_stats}
\end{table}

\paragraph{Dataset construction.}
For each level in the regular and extra splits, we construct two transition datasets.
The \textit{solution transition dataset} is obtained from successful human solution trajectories and emphasizes the transitions that arise along task-completing behavior.
The \textit{coverage transition dataset} is obtained by running BFS on each level and recording canonical transitions until either the reachable frontier is exhausted or \(100{,}000\) canonical transitions have been collected for that level.
Both datasets are stored as transition archives paired with state archives.
Within each level-specific dataset, we canonicalize states and remove duplicate transitions that share the same \((s,a,s')\) triple.
This ensures that repeated visits to the same local transition effect do not inflate the archived dataset and that the \(100{,}000\)-transition cap for coverage is applied to canonical transitions rather than raw observations.
Aggregate dataset statistics are summarized in \autoref{tab:appendix_dataset_stats}.
The class-reduced subsets used for balanced accuracy are derived from these full transition datasets by the heuristic procedure described in \appref{sec:appendix_heuristic_dynamics}.

\paragraph{Evaluation protocol and cost measurement.}
After training, we evaluate the selected executable world model on the full and class-reduced variants of the solution transition dataset and coverage transition dataset for each split.
All Acc. is measured on the full transition dataset, whereas Balanced Acc. is measured on the corresponding class-reduced subset derived by heuristic dynamics discovery.
Our primary metric is one-step transition accuracy.
For each archived transition \((s,a,s')\), a prediction is counted as correct only if the program executes successfully on \((s,a)\) and its predicted next state exactly matches the archived next state after canonicalization.
Compile failures, contract violations, missing predictions, and runtime errors are all counted as incorrect predictions.
We report regular-set and extra-set accuracy separately by aggregating the corresponding archived datasets over levels within each split.
For LLM-based methods, we report cumulative training-time Calls, token composition, and relative weighted token cost; Calls alone is not a total-cost measure (\appref{sec:appendix_cost}).

\paragraph{Compute resources.}
Experiments were run on a workstation with an Intel Xeon Gold 6226R CPU at 2.90GHz, 128GB RAM, and two NVIDIA GeForce RTX 3090 GPUs.
The operating system was Ubuntu 20.04.6 LTS with Linux kernel 5.4.0-216-generic.
Elapsed-time measurements for the five-run experiments are reported in \appref{sec:appendix_cost}. External LLM latency affects wall-clock time; token usage and local computation are therefore reported separately.

\paragraph{Embedding architecture and optimization.} In Baba, each object is a 58-dimensional token: a 4-dimensional one-hot type, a 45-dimensional one-hot word, a 5-dimensional one-hot direction, and four dimensions for normalized position and grid scale. We project tokens to 384 dimensions and encode their relations with four six-head Set Transformer blocks. Object order has no meaning, so we use no positional encoding. A query combining a learnable seed and a 384-dimensional action embedding pools the token set; an MLP and L2 normalization produce $h_\theta(s,a)\in\mathbb R^{384}$. The encoder has approximately 6.49 million trainable parameters. We use a learning rate of $10^{-4}$, supervised-contrastive temperature $0.2$, and two-GPU distributed data parallel training with global batch size 4,096. Every 4,096 added transitions trigger 30 optimization updates. Each active class is represented by up to 8 prototypes for frontier scoring. These object-token dimensions describe the Baba encoder, not ALFWorld's distinct symbolic state representation.

\paragraph{Update and stopping rules.}
Unless otherwise noted, the Baba online discovery runs use the same pipeline and differ only in the component being ablated. The ALFWorld interaction budget is specified in \autoref{sec:exp-setup}.
Each run terminates when either the global LLM call budget of \(100\) is exhausted, a single program-update iteration uses \(15\) LLM calls without an accepted update, or no candidate update has been accepted for \(300{,}000\) global interaction steps.
Inductive program update uses a per-target retry cap of \(15\) rejected candidate updates.
When a rejected candidate update causes previously explained transitions to become unexplained, we sample additional transition evidence by selecting up to \(3\) affected classes uniformly at random and then selecting one broken transition uniformly at random from each selected class.
Candidate updates are accepted only if they resolve the target failure and introduce no compile errors, runtime errors, or explanation losses on the full set of previously explained transitions in \(D_t\).
For Explorer representation learning, we include a hypothesis class in the active class set only after it has at least \(8\) assigned transitions, and the learned state-action embedding has dimension \(384\).

Across LLM-based runs, we keep the prompting and decoding configuration fixed unless the experiment explicitly studies backbone variation.
In particular, we use a medium reasoning or thinking setting together with a maximum output budget of \(32{,}768\) tokens.
In the backbone comparison, we change only the underlying LLM while keeping the update procedure, exploration strategy, prompts, and evaluation protocol unchanged.

\subsection{ALFWorld Experiments}
\label{sec:appendix_alfworld}
\paragraph{Setting and metric.}\leavevmode
The protocol in \autoref{sec:exp-setup} includes all six household task types. Evaluation episodes are not used during learning, and the metric is exact next-state agreement; compile failures and runtime errors count as incorrect predictions. Alice uses the fixed coefficient $\lambda_{\mathcal C}=0.35$. The main comparison trains both methods without a curriculum and reports means and standard deviations over five runs.

\paragraph{Comparison and curriculum.}
The curriculum comparison in \autoref{tab:alfworld_results} uses all six task types while preserving WorldCoder's original task order. It uses the same unseen-episode evaluation and call caps as the no-curriculum comparison.

WorldCoder's original ALFWorld results report average episode reward (task success) over three seeds with an MCTS planner and a BM25 heuristic. Its curriculum covers four of the six task types, and its learning-performance evaluation has no separate held-out split. Our six-task, unseen-episode transition evaluation therefore differs in metric, task coverage, and evaluation protocol. High planning success does not require exact predictions on every transition outside the successful plans, so the two reported accuracies should not be directly compared.

\paragraph{Token cost.}
\autoref{tab:appendix_alfworld_cost} reports token usage under the same accounting convention as \appref{sec:appendix_cost}. Alice's weighted cost is less than half that of WorldCoder in this setting, alongside higher prediction accuracy. This result is specific to ALFWorld and does not imply a cost advantage in Wonderland online learning. The curriculum diagnostic is not included in this cost table.

\begin{table}[!htbp]
\centering\small
\setlength{\tabcolsep}{2pt}
\caption{ALFWorld token accounting. Token counts are in millions; weighted cost is measured in millions of input-token-equivalent units. Five-run mean $\pm$ standard deviation.}
\label{tab:appendix_alfworld_cost}
\begin{tabular*}{\textwidth}{@{\hspace{4pt}\extracolsep{\fill}}lcc@{\hspace{4pt}}}
\toprule
Quantity & \textbf{Alice} & \textbf{WorldCoder} \\
\midrule
Input tokens & $1.332\pm0.467$ & $2.861\pm0.247$ \\
Reasoning tokens & $0.010\pm0.004$ & $0.057\pm0.009$ \\
Output tokens & $0.029\pm0.007$ & $0.109\pm0.029$ \\
Total tokens & $1.371\pm0.476$ & $3.026\pm0.258$ \\
Weighted cost & $1.564\pm0.525$ & $3.856\pm0.372$ \\
\bottomrule
\end{tabular*}
\end{table}

\subsection{Heuristic Dynamics Discovery for Class-Reduced Subsets}
\label{sec:appendix_heuristic_dynamics}

\paragraph{Algorithm.}
The heuristic dynamics discovery procedure is applied to the archived transition datasets after collection.
For each transition, we first reconstruct the source state and next state from the paired state archive.
Each state is represented as a multiset of object tokens of the form \((\texttt{type}, \texttt{word}, x, y, \texttt{direction})\), together with the grid size and termination flags.
We then compute a canonical state-difference signature that summarizes how the object multiset changes across the transition.

This signature records three kinds of object-level effects: moved objects, removed objects, and added objects.
For moved objects, we match objects by \texttt{type}, \texttt{word}, and \texttt{direction}, and record only the displacement \((\Delta x, \Delta y)\) together with the object identity and multiplicity.
For objects that disappear or appear without a matched counterpart, we record removal or addition counts by object identity.
The signature also records changes in grid size, termination, or truncation when they occur.

Two transitions are assigned to the same heuristic class if and only if they have the same action and the same canonical state-difference signature.
As a result, transitions that induce the same local dynamical effect but occur at different absolute positions or in different levels are collapsed into the same heuristic class.
This heuristic is not intended to recover the true latent dynamics of the environment.
Rather, it provides a simple way to remove large amounts of redundant transition evidence while preserving diversity over distinct observed transition effects.

\paragraph{Subset construction.}
Given the heuristic class assignments above, we construct a class-reduced subset by selecting one representative transition from each heuristic class.
For each heuristic class, we select one representative transition by uniform random sampling with a fixed seed.
The resulting subset therefore contains exactly one transition per heuristic class.

We apply this procedure separately to the solution transition dataset and the coverage transition dataset for each split.
The resulting class-reduced subsets are used to compute Balanced Acc. for the solution transition dataset and coverage transition dataset.
In our archived datasets, this reduces the train/test solution transition sets from \(2{,}476/1{,}126\) transitions to \(595/293\) representatives and the train/test coverage transition sets from \(3{,}200{,}000/800{,}000\) transitions to \(7{,}232/2{,}073\) representatives.
These subsets are not intended to be harder benchmarks than the full datasets.
Rather, they provide a complementary view that discounts repeated instances of the same local transition effect and emphasizes coverage over distinct heuristic dynamics classes.

\subsection{Computational Cost and Scaling}
\label{sec:appendix_cost}
\paragraph{API accounting.}\leavevmode
Calls alone do not measure total API cost. We record input, reasoning, and output tokens separately for each run. Under the pricing ratio used in these experiments, reasoning and output tokens each cost six times as much as input tokens. We therefore compute $\mathrm{Cost}=\mathrm{Input}+6(\mathrm{Reasoning}+\mathrm{Output})$ per run and then report the mean and standard deviation. This is a relative token-price measure, not a dollar amount or a measure of local compute. Totals and costs are computed before rounding.

\begin{table}[!htbp]
\centering\small
\setlength{\tabcolsep}{2pt}
\caption{Wonderland token accounting. Token counts are in millions; weighted cost is measured in millions of input-token-equivalent units. Five-run mean $\pm$ standard deviation.}
\label{tab:appendix_token_cost}
\begin{tabular*}{\textwidth}{@{\hspace{4pt}\extracolsep{\fill}}llccc@{\hspace{4pt}}}
\toprule
Setting & Method & Input & Reasoning & Output \\
\midrule
Online & \textbf{Alice} & $2.867\pm0.531$ & $0.096\pm0.015$ & $0.097\pm0.026$ \\
Online & \textbf{WorldCoder} & $1.335\pm0.778$ & $0.053\pm0.024$ & $0.096\pm0.015$ \\
Offline & \textbf{Alice} & $3.020\pm0.429$ & $0.115\pm0.019$ & $0.132\pm0.073$ \\
Offline & \textbf{GIF-MCTS} & $2.144\pm0.145$ & $0.070\pm0.046$ & $0.230\pm0.069$ \\
Offline & \textbf{CWM} & $2.289\pm0.160$ & $0.167\pm0.013$ & $0.245\pm0.032$ \\
\bottomrule
\end{tabular*}
\par\medskip
\begin{tabular*}{\textwidth}{@{\hspace{4pt}\extracolsep{\fill}}llcc@{\hspace{4pt}}}
\toprule
Setting & Method & Total tokens & Weighted cost \\
\midrule
Online & \textbf{Alice} & $3.060\pm0.555$ & $4.025\pm0.689$ \\
Online & \textbf{WorldCoder} & $1.485\pm0.764$ & $2.232\pm0.707$ \\
Offline & \textbf{Alice} & $3.267\pm0.497$ & $4.500\pm0.867$ \\
Offline & \textbf{GIF-MCTS} & $2.444\pm0.126$ & $3.945\pm0.079$ \\
Offline & \textbf{CWM} & $2.701\pm0.187$ & $4.763\pm0.319$ \\
\bottomrule
\end{tabular*}
\end{table}

In Wonderland online learning, Alice incurs a higher weighted token cost than WorldCoder while achieving higher accuracy. WorldCoder's lower expenditure is real, although it partly reflects termination at $57.6$ calls on average. In the offline comparison, Alice's weighted cost lies between those of GIF-MCTS and CWM. Thus, lower call counts should not be read as a general total-cost advantage.

\paragraph{Embedding overhead.}
\autoref{tab:appendix_embedding_cost} separates frontier scoring, prototype refresh, and embedding training in the five online runs. Together these stages average $1.26\pm0.93$ hours, or $21.33\pm10.66\%$ of total wall-clock time. These are stage elapsed times, not GPU-hours. Percentages are computed within each run before averaging, so a ratio of the reported means need not equal the mean percentage. We do not infer a baseline wall-clock advantage because external LLM latency affects total elapsed time.

\begin{table}[!htbp]
\centering\small
\setlength{\tabcolsep}{2pt}
\caption{Embedding-related elapsed time and share of total wall-clock time in online Wonderland runs. Five-run mean $\pm$ standard deviation.}
\label{tab:appendix_embedding_cost}
\begin{tabular*}{\textwidth}{@{\hspace{4pt}\extracolsep{\fill}}lcc@{\hspace{4pt}}}
\toprule
Stage & Time (hours) & Share (\%) \\
\midrule
Frontier scoring & $0.39\pm0.31$ & $6.39\pm3.86$ \\
Prototype refresh & $0.19\pm0.15$ & $3.23\pm1.73$ \\
Embedding training & $0.68\pm0.48$ & $11.72\pm5.08$ \\
Total & $1.26\pm0.93$ & $21.33\pm10.66$ \\
\bottomrule
\end{tabular*}
\end{table}

\paragraph{Acceptance-check scaling.}
We group the 470 candidate checks from the five online Wonderland runs by the size of the accumulated explained set (\autoref{tab:appendix_accept_scaling}). The check median is $2.99$ seconds overall and $85.00$ seconds in the largest bucket; candidate program complexity also increases during learning. Full-set checking is therefore a non-trivial local CPU cost at large evidence-set sizes. It is performed when evaluating a proposed update, rather than at every interaction, and consumes no additional API tokens. The largest bucket contains 53 of the 470 checks. Meanwhile, median tokens per call rise from 8,786 to 36,354 across the smallest and largest buckets. The update prompt contains at most three selected broken examples, so it does not include the entire growing evidence set; program size and example length can still increase context cost.

\begin{table}[!htbp]
\centering\small
\setlength{\tabcolsep}{2pt}
\caption{Acceptance-check scaling. Each entry is median (90th percentile); $n$ counts candidate checks, not independent runs.}
\label{tab:appendix_accept_scaling}
\begin{tabular*}{\textwidth}{@{\hspace{4pt}\extracolsep{\fill}}lrccc@{\hspace{4pt}}}
\toprule
Explained-set size & $n$ & Check time (s) & AST nodes & Tokens / call \\
\midrule
$[0,10)$ & 14 & $0.01\ (0.02)$ & $1{,}065.5\ (1{,}295.9)$ & $8{,}786\ (14{,}952)$ \\
$[10,100)$ & 35 & $0.04\ (0.07)$ & $1{,}632.0\ (2{,}251.4)$ & $17{,}880\ (35{,}576)$ \\
$[100,1{,}000)$ & 81 & $0.17\ (0.23)$ & $3{,}178.0\ (4{,}043.0)$ & $23{,}964\ (58{,}948)$ \\
$[1{,}000,10{,}000)$ & 89 & $0.63\ (1.72)$ & $4{,}371.0\ (5{,}246.6)$ & $20{,}676\ (57{,}025)$ \\
$[10{,}000,10^5)$ & 198 & $5.84\ (16.41)$ & $6{,}138.0\ (8{,}544.2)$ & $32{,}163\ (59{,}127)$ \\
$[10^5,10^6)$ & 53 & $85.00\ (244.60)$ & $7{,}507.0\ (10{,}064.8)$ & $36{,}354\ (72{,}574)$ \\
All checks & 470 & $2.99\ (35.47)$ & $5{,}126.0\ (8{,}513.3)$ & $25{,}833\ (60{,}307)$ \\
\bottomrule
\end{tabular*}
\end{table}

\clearpage
\input{checklist.tex}

\end{document}

%% file: checklist.tex
\section*{NeurIPS Paper Checklist}

\begin{enumerate}

\item {\bf Claims}
    \item[] Question: Do the main claims made in the abstract and introduction accurately reflect the paper's contributions and scope?
    \item[] Answer: \answerYes{}
    \item[] Justification: The abstract and introduction describe a general method for online executable world-model learning and a controlled probe of lexical-prior dependence. Five-run Baba and ALFWorld results support the stated scope. The ablations report overall trends as class structure and exploration signals are added, supported by trend tests.

\item {\bf Limitations}
    \item[] Question: Does the paper discuss the limitations of the work performed by the authors?
    \item[] Answer: \answerYes{}
    \item[] Justification: The conclusion limits the method to deterministic, fully observable symbolic settings with exact execution checks. \appref{sec:appendix_cost} discusses full-set checking and embedding overhead; qualitative examples show concrete prediction failures. ALFWorld does not establish applicability to raw visual or partially observed inputs.

\item {\bf Theory assumptions and proofs}
    \item[] Question: For each theoretical result, does the paper provide the full set of assumptions and a complete (and correct) proof?
    \item[] Answer: \answerNA{}
    \item[] Justification: The paper does not present formal theoretical results requiring proofs. The formal setup and information-theoretic view are used to define the problem and motivate the frontier score, not to state theorems.

    \item {\bf Experimental result reproducibility}
    \item[] Question: Does the paper fully disclose all the information needed to reproduce the main experimental results of the paper to the extent that it affects the main claims and/or conclusions of the paper (regardless of whether the code and data are provided or not)?
    \item[] Answer: \answerYes{}
    \item[] Justification: The paper specifies the benchmark construction, train/test level splits, evaluation datasets, metrics, LLM-call budgets, termination criteria, prompt template, and implementation details needed to reproduce the main experiments. Additional dataset statistics and evaluation protocols are provided in~\appref{sec:appendix_experimental_details}.

\item {\bf Open access to data and code}
    \item[] Question: Does the paper provide open access to the data and code, with sufficient instructions to faithfully reproduce the main experimental results, as described in supplemental material?
    \item[] Answer: \answerYes{}
    \item[] Justification: We provide the source code, released transition datasets, data-generation and evaluation scripts, and reproduction instructions in the supplementary material. The paper documents the dataset construction, level splits, metrics, LLM-call budgets, and evaluation protocol needed to reproduce the main results.

\item {\bf Experimental setting/details}
    \item[] Question: Does the paper specify all the training and test details (e.g., data splits, hyperparameters, how they were chosen, type of optimizer) necessary to understand the results?
    \item[] Answer: \answerYes{}
    \item[] Justification: \autoref{sec:exp-setup} and \appref{sec:appendix_experimental_details} describe the Baba splits, metrics, budgets, embedding architecture, training schedule, and decoding settings. \appref{sec:appendix_revised_results} reports the coefficient sweep and default setting; \appref{sec:appendix_alfworld} gives the ALFWorld protocol and curriculum diagnostic.

\item {\bf Experiment statistical significance}
    \item[] Question: Does the paper report error bars suitably and correctly defined or other appropriate information about the statistical significance of the experiments?
    \item[] Answer: \answerYes{}
    \item[] Justification: The main comparisons and ablations report means and standard deviations over five independent runs, with the run as the unit of analysis. \appref{sec:appendix_revised_results} specifies exact pairwise permutation and trend tests and their Holm corrections. Individual ablation comparisons are not significant after correction; the three ordered trends are. Additional analyses state their representative-run configuration.

\item {\bf Experiments compute resources}
    \item[] Question: For each experiment, does the paper provide sufficient information on the computer resources (type of compute workers, memory, time of execution) needed to reproduce the experiments?
    \item[] Answer: \answerYes{}
    \item[] Justification: \appref{sec:appendix_experimental_details} specifies hardware and embedding training settings. \appref{sec:appendix_cost} reports token usage, relative weighted cost, embedding-stage elapsed times, and acceptance-check scaling; \appref{sec:appendix_alfworld} separately reports ALFWorld costs.
    
\item {\bf Code of ethics}
    \item[] Question: Does the research conducted in the paper conform, in every respect, with the NeurIPS Code of Ethics \url{https://neurips.cc/public/EthicsGuidelines}?
    \item[] Answer: \answerYes{}
    \item[] Justification: The research uses controlled symbolic game environments and archived transition data, and does not involve sensitive personal data, human-subject experimentation, or deployment on real users. The work is intended as foundational research on executable world-model learning.

\item {\bf Broader impacts}
    \item[] Question: Does the paper discuss both potential positive societal impacts and negative societal impacts of the work performed?
    \item[] Answer: \answerNA{}
    \item[] Justification: This work is foundational research on online executable world-model learning in controlled symbolic game environments, with no deployment, human subjects, personal data, or application-specific decision making. We therefore do not identify direct societal impacts beyond the general indirect effects of progress in model-based AI systems.
    
\item {\bf Safeguards}
    \item[] Question: Does the paper describe safeguards that have been put in place for responsible release of data or models that have a high risk for misuse (e.g., pre-trained language models, image generators, or scraped datasets)?
    \item[] Answer: \answerNA{}
    \item[] Justification: The released assets consist of source code and controlled symbolic game transition datasets, not high-risk pretrained models, generative models, scraped web data, or sensitive personal data. We therefore do not identify misuse risks that require special release safeguards.

\item {\bf Licenses for existing assets}
    \item[] Question: Are the creators or original owners of assets (e.g., code, data, models), used in the paper, properly credited and are the license and terms of use explicitly mentioned and properly respected?
    \item[] Answer: \answerYes{}
    \item[] Justification: We credit \textit{Baba Is You} as the source of the underlying game assets and level/map data used in the benchmark. We obtained permission from the game's developer to use these assets and map data for this research, and the submitted code/data package includes the corresponding attribution and usage information.

\item {\bf New assets}
    \item[] Question: Are new assets introduced in the paper well documented and is the documentation provided alongside the assets?
    \item[] Answer: \answerYes{}
    \item[] Justification: The submitted assets include the \textit{Baba in Wonderland} benchmark code and transition datasets. The paper documents the remapping, state/action representation, level splits, dataset construction procedure, dataset statistics, and evaluation protocol, with additional reproduction details provided in the code package.

\item {\bf Crowdsourcing and research with human subjects}
    \item[] Question: For crowdsourcing experiments and research with human subjects, does the paper include the full text of instructions given to participants and screenshots, if applicable, as well as details about compensation (if any)? 
    \item[] Answer: \answerNA{}
    \item[] Justification: The paper does not involve crowdsourcing or human-subject experiments. Human-play solution trajectories are used only as archived gameplay transition data and do not include participant studies, compensation, or personal information.

\item {\bf Institutional review board (IRB) approvals or equivalent for research with human subjects}
    \item[] Question: Does the paper describe potential risks incurred by study participants, whether such risks were disclosed to the subjects, and whether Institutional Review Board (IRB) approvals (or an equivalent approval/review based on the requirements of your country or institution) were obtained?
    \item[] Answer: \answerNA{}
    \item[] Justification: The work does not involve crowdsourcing or human-subject research requiring IRB approval or equivalent review. No personal data or participant intervention is used.

\item {\bf Declaration of LLM usage}
    \item[] Question: Does the paper describe the usage of LLMs if it is an important, original, or non-standard component of the core methods in this research? Note that if the LLM is used only for writing, editing, or formatting purposes and does \emph{not} impact the core methodology, scientific rigor, or originality of the research, declaration is not required.
    %this research? 
    \item[] Answer: \answerYes{}
    \item[] Justification: LLMs are a core component of Alice's Programmer, and the paper describes their use for executable program updates, reports LLM-call budgets, includes the program-update prompt template, and evaluates backbone variation in the appendix.

\end{enumerate}